\pdfoutput=1

\documentclass[11pt]{article}
\usepackage{booktabs} 
\usepackage{acl}
\usepackage{graphicx}
\usepackage[most]{tcolorbox}
\usepackage{times}
\usepackage{latexsym}
\usepackage{tcolorbox}
\usepackage{booktabs} 
\usepackage{capt-of}
\usepackage{placeins}
\usepackage{cuted}
\usepackage[T1]{fontenc}
\usepackage{multicol}
\usepackage[utf8]{inputenc}

\usepackage{microtype}

\title{Learning to Plan by Updating Natural Language}

\author{
Yiduo Guo$^{1,}$,~~Yaobo Liang$^{2}$,~~Chenfei Wu$^{2}$,~~Wenshan Wu$^{2}$,~~Dongyan Zhao$^{1}$,~~Nan Duan$^{2}$\\ 
$^1$Wangxuan Institute of Computer Technology, Peking University\\
$^2$Microsoft Research Asia\\
\texttt{yiduo@stu.pku.edu.cn,zhaodongyan@pku.edu.cn}, \\\texttt{\{yaobo.liang, chenfei.wu, wenshan.wu, nanduan\}@microsoft.com}\\
}

\begin{document}
\maketitle
\begin{abstract}
Large Language Models (LLMs) have shown remarkable performance in various
basic natural language tasks. For completing the complex task, we still need a plan for the task to guide LLMs to generate the specific solutions step by step. 
LLMs can directly generate task plans, but these plans may still contain factual errors or are incomplete. A high-quality task plan contains correct step-by-step solutions for solving all situations and behavioral instructions for avoiding mistakes. To obtain it, we
propose the Learning to Plan method, which involves two phases: (1) In the first learning task plan phase, it iteratively updates the task plan with new step-by-step solutions and behavioral instructions, which are obtained by prompting LLMs to derive from training error feedback. 
(2) In the subsequent test phase, the LLM uses the learned task plan to guide
the inference of LLM on the test set. We demonstrate the effectiveness of our method on the five different reasoning type tasks (8 datasets).
Further, our analysis experiment shows that the task plan learned by one LLM can directly guide another LLM to improve its performance, which reveals a new transfer learning paradigm.\footnote{We release the code at \url{https://github.com/Eureka6174/LearnNLPlan}} 
\end{abstract}

\section{Introduction}

\label{sec.intro}
\vspace{-1mm}
Large Language Models (LLMs), such as ChatGPT and GPT-4 ~\cite{openai2023gpt4}, have recently achieved strong zero-shot/few-shot performance on various natural language tasks, such as generating passages~\cite{bang2023multitask}, generating code~\cite{liu2023comprehensive}, and solving grade school math problems~\cite{qin2023chatgpt}. LLMs can further learn new basic abilities by connecting them with millions of APIs like TaskMatrix.AI ~\cite{liang2023taskmatrix, wu2023visual} or new tools like ToolFormer~\cite{schick2023toolformer}.
However, LLMs still struggle to complete complex tasks, such as writing a long novel~\cite{yang2022doc}, coding for a large project~\cite{orlandoassessing}, and solving complex math problems~\cite{frieder2023mathematical}. This indicates that knowing every basic step/capability is insufficient to complete complex tasks - we still require a task plan consisting of step-by-step solutions and behavioral instructions to solve the complex task. 
\begin{table}
\centering
\begin{tabular}{|p{0.95\linewidth}|}
\toprule 
\small{\textbf{Task instruction (prompt):} Calculating the sin/cos value of an angle based on the length of legs and other conditions in a triangle.} \\
\hline
\small{\textbf{Input:} In a triangle ABC, AC is 8 units, and BC is 15 units, what is cos<ABC?} \\
\hline
\small{\textbf{Label:}$\frac{15}{17}$} \\
\hline
\small{\textbf{Zero-shot chain of thought:}~Let's think step by step. 

\textbf{Corresponding solution of the input:} 
Apply the cosine rule: $\cos(\angle ABC) = \frac{8^2 + 15^2 - 15^2}{2 \cdot 8 \cdot 15}= \frac{64}{240}$

Simplify: $\cos(\angle ABC) = \frac{4}{15}$}\\
\hline
\small{\textbf{Task plan learned by our method}: There are two solutions for finding the value of an angle in a triangle. Solution 1 involves using the Law of Cosines to calculate $\cos(\angle ABC)$.} \\\small{Solution 2 involves using the Pythagorean Theorem to find the length of the third side of the triangle and then using the definition of sine to calculate $\sin(\angle ABC)$. To find the length of the hypotenuse of a right triangle using the Pythagorean Theorem, use the formula $(\overline{AC})^2 + (\overline{BC})^2 = (\overline{AB})^2$. python functions can also be used to calculate these values.

\textbf{Corresponding solution of the input}: Using the Pythagorean theorem, we can have $AB=\sqrt{8^2+15^2}=17$. Then, based on the definition of cosines, we can have $cos<ABC=\frac{15}{17}$}\\

\bottomrule
\end{tabular}
\caption{We list the typical elements of a task.}
\label{tab.illustration}
\vspace{-5pt}
\end{table}
\normalsize
We can employ human experts to write high-quality task plans for solving complex tasks. However, writing a task plan by humans is hard and expensive. To automatize this, We can prompt the LLM to generate the task plan and then use it to guide the test inference. However, LLMs may generate errors or incomplete steps in the plan without the supervision of feedback.  To address the problem, in this paper, we propose the learning to plan method, which automatically learns the task plan containing step-by-step solutions and behavioral instructions by iteratively prompting the LLM to solve the errors. {\color{black} We present the task plan learned by our method in Table~\ref{tab.illustration}. In contrast to the task instruction, our task plan contains multiple step-by-step solutions. In comparison to the specific solution, our task plan comprises general solutions that can guide the LLM to solve all task samples.

Specifically, we first collect the wrong samples that can not be solved with the guidance of the current task plan. This step is called the \textbf{finding wrong samples with current task plan} step.  ~Next, in the \textbf{computing plan update} step, we prompt the LLM to generate multiple new step-by-step solutions for solving wrong samples and samples similar to them. We choose the solution that achieves the best performance on a validation set as the plan update. Text compression is also considered for deduplication. ~In the \textbf{updating plan} step, we update the task plan by directly adding the best new solution (text) to the original task plan (text). 
We repeat the learning process until we can not find a new solution to improve the task plan's performance. In the test phase, we add the learned task plan into the prompt to guide the LLM. }

Experiments on 8 reasoning datasets with five reasoning types: mathematical reasoning, causal reasoning, logical reasoning, symbolic reasoning
task, and combinatorial reasoning show that our method improves the LLM's performance of these tasks significantly. For example, our method's average performance of 10 tasks from the AMPS mathematical dataset~\cite{hendrycksmath2021} outperforms the performance directly measured in the zero-shot/few-shot chain of thought setting by 18.3$\%$/7$\%$ respectively (Table~\ref{tab:Khan}).
We also find that our method can improve the zero-shot performance of the newest open-source LLMs (e.g., Vicuna v1.5). That verifies our method is a general algorithm for both closed-source and open-source LLMs. 
Furthermore, we find that the learned plan by ChatGPT can directly guide GPT-4 to improve GPT-4's performance. This means our learned task plan in natural language can transfer between LLM A and another LLM B. 

The paper's main conclusions are as follows:

(1) To solve the complex task, we propose the learning to plan method that prompts LLMs to learn and update the task plan from the errors iteratively. The task plan can guide the LLM in the test phase.

(2) We verify the effectiveness of our method on 8 reasoning datasets involving 5 different reasoning types. With the guidance of the task plan learned by our method, the LLM improves its zero-shot/few-shot performance markedly. Our method works well for both closed-source and open-source LLMs.

(3) Further, we find that the task plan learned by one LLM (e.g., ChatGPT) can guide another LLM (e.g., GPT-4) to improve its performance, which implies that our learned task plan can transfer between different LLMs as a general natural language task experience.

\section{Related Work}
\label{sec.related}
\vspace{-1mm}
\textbf{Large Language Models} (e.g., GPT3~\cite{brown2020language}, ChatGPT, OPT~\cite{zhang2022opt}, and LLaMA~\cite{touvron2023llama})recently achieve huge progress. They can generate simple code~\cite{chen2021evaluating}, solve simple math problem\cite{bang2023multitask}, and write the draft of mathematical proof\cite{jiang2022draft}.
 Further, as a text understanding and generation module, they can connect other APIs ~\cite{liang2023taskmatrix}, models ~\cite{wu2023visual}, and tools~\cite{schick2023toolformer} to expand basic abilities for solving more complex tasks. But it's still challenging for LLMs to solve complex tasks like writing a novel. 
Our paper focuses on the learning problem for the deployed large language model to solve complex tasks. 

\textbf{Prompt Engineering}~\cite{liu2023pre,wei2022chain,kojima2022large} further improves the performance of LLMs by inserting a prompt into the input.  Discrete prompts consist of natural language phrases.~\cite{jiang2020can} uses the mining-based methods to extract Middle words and words in the dependency path as the potential discrete prompts. Prompt-paraphrasing methods~\cite{jiang2020can,haviv2021bertese} create the potential prompts by paraphrasing the original prompt into a set of other prompts.~\cite{gao2020making} uses pre-trained T5 to generate discrete prompts. However, discrete prompts can not update the previous prompt to make the prompt better and do not induce the task information into their method.
Continuous prompts~\cite{li2021prefix,zhong2021factual,liu2021gpt} perform
prompting directly in the embedding space of the model. Their templates have their own parameters that can be tuned in the training dataset. But humans can not directly understand them.

\textbf{Task Planning Methods} can be divided into two main approaches: One approach prompts the LLM to break the problem down into sub-problems without feedback. Chain of Thought (CoT) prompting and Zero-shot CoT~\cite{wei2022chain,kojima2022large} generate reasoning intermediate processes for complex
problems by prompting the model to "think step by step. Least-to-most prompting~\cite{zhou2022least} prompts LLMs to decouple a complex problem into a list of sub-problems and solve them sequentially. 
Another approach uses feedback to decouple complex tasks step by step. Self-refine method and related methods~\cite{madaan2023self,saunders2022self,yao2023retroformer,zheng2023progressive} and Tree of thoughts~\cite{yao2023tree} improve LLMs' specific solution iteratively by utilizing the feedback provided by LLM itself. React~\cite{yao2022react}, Progprompt~\cite{singh2023progprompt}, and Inner Monologue ~\cite{huang2022inner} uses environmental feedback to generate specific action plan step by step. APE~\cite{zhou2022large} and Instruction Induction~\cite{honovich2022instruction} update task instruction by iteratively generating instruction candidates and selecting the one with the best validation performance. 
\section{Method}
\subsection{Task Plan and Self-Plan Method}
To solve the complex task, we need a task plan in natural language that consists of step-by-step solutions and behavioral instructions to guide the LLM. For example, to calculate the roots of the quadratic equation, the plan can be made of the step-by-step solution based on the quadratic formula and instruction for taking care of the no real roots situation. The task plan should be (1) generally applicable to any sample from the task and (2) easily comprehensible by humans, which allows humans to read, edit, and add new solutions or instructions to improve the quality of the plan. 

To get a high-quality task plan, we can employ human experts to write the step-by-step solutions for the task. However, the labor cost is high when we need to complete a huge amount of tasks. Another way is to prompt the LLM to generate the task plan. Specifically, we ask the LLM to generate a task plan that includes general step-by-step solutions to solve the task (see Figure~\ref{prompt.1} in Appendix~\ref{appendx.self-plan}). Next, we use the generated plan to guide the inference of the test sample $x$ of the task. This way utilizes the knowledge of the pre-trained LLM. However, as we showed in the experiment section, the generated task plan may have many factual errors and incomplete steps. To automatically obtain a high-quality plan and reduce factual errors, we propose the {learning to plan} method. The learned plan by our method boosts the LLM's complex task performance markedly. 

\subsection{Learning to Plan}
\begin{figure*}[htp]
  \centering
  \includegraphics[width=1\textwidth]{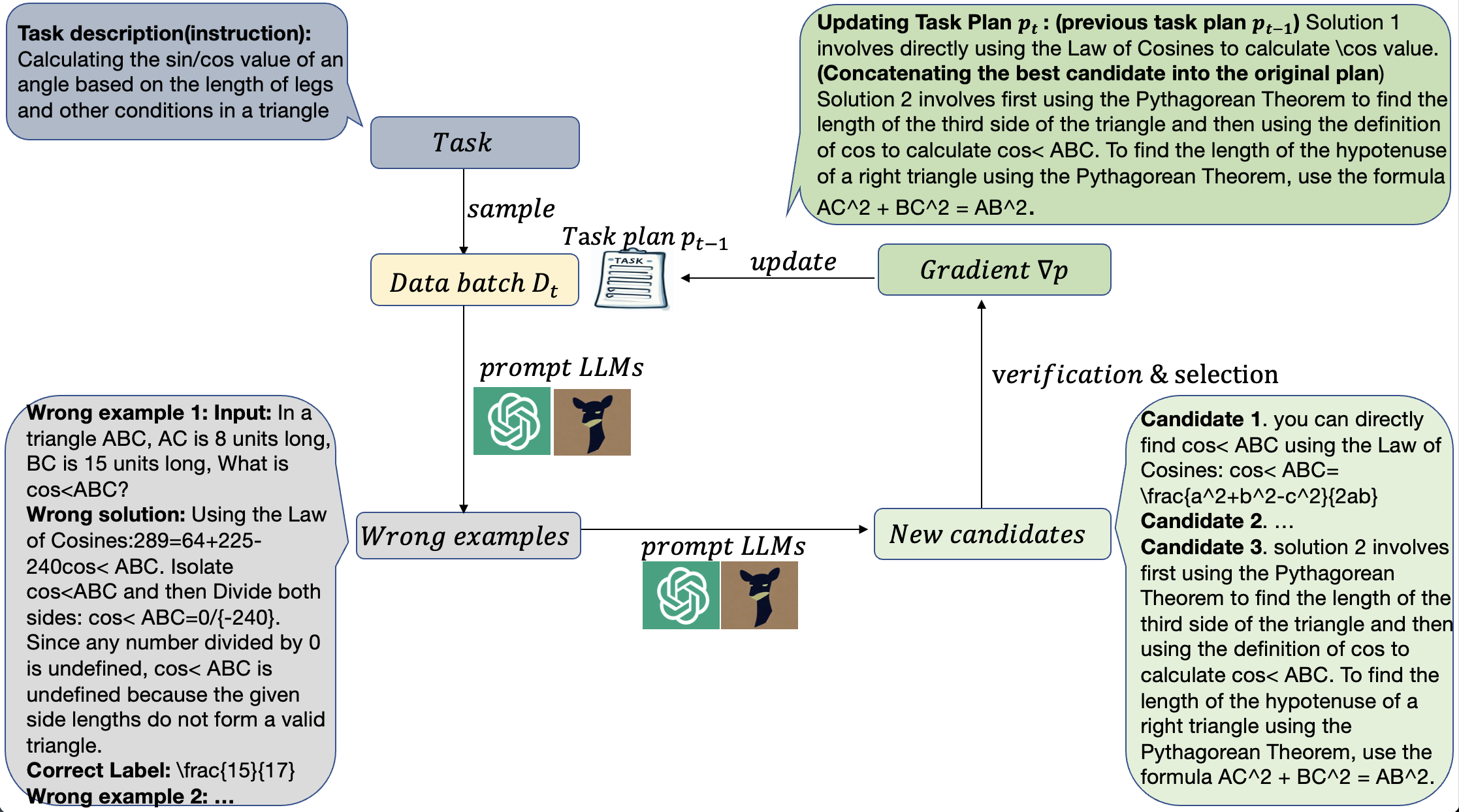}
  \caption{The whole learning process of our learning to plan method and examples in the Rounded Rectangle Annotations.}
  \label{fig:LP}
\vspace{-5pt}
\end{figure*}
In this section, we introduce the learning to plan method that learns and updates the task plan $p$ based on the error feedback. Similar to the traditional deep learning setting, we divide the training set $D$ into equal-size training data batches. We consider one training epoch to be completed once the model has processed all data batches in one pass. Our method's learning procedure for each training iteration consists of three steps: (1) finding errors that can not be solved with the current task plan (2) prompting the LLM to generate new step-by-step solutions and behavioral instructions for solving the errors and verifying the effectiveness of new solutions and instructions (3) updating the task plan with new solutions and instructions. We illustrate the whole learning procedure in Figure~\ref{fig:LP}.

\subsubsection{Finding Wrong Samples With Current Task Plan}  During the $t$-th training iteration, given the data batch $D_t=\{{x_i},y_i\}_{i=1}^{d}$ and the current task plan $p_{t-1}$, we collect the wrong samples that can not be solved with the guidance of the current task plan $p_{t-1}$. Note that we set the initial task plan $p_0$ as the empty string for the first iteration. Specifically, we first prompt the LLM to make prediction $H(x_i,p)$ for each data point in the batch with the guidance of task plan $p_{t-1}$. Then we collect all wrong predictions, their inputs, and their correct labels to construct the wrong example set $\{{x_i}, H(x_i, p_{t-1}),y_i\}_{i=1}^{d^{'}}$ by comparing the prediction with the true label. We illustrate this step on the left side of Figure~\ref{fig:LP}. 

\subsubsection{Computing Plan Update}
We update the task plan by adding the plan update consisting of new step-by-step solutions and instructions $\nabla p$. 
\label{sec.revision}
To obtain $\nabla p$, we first prompt the LLM to generate multiple plan update candidates that can correctly solve the wrong examples. Specifically, (1) we randomly choose $m$ wrong examples from the wrong example set. Then we use the $m$ wrong examples, previous task plan $p_{t-1}$, and a new solution generation prompt (see Figure~\ref{prompt.2}) as input to prompt the LLM to generate the new step-by-step solutions to solve the errors. (2) we repeat (1) $K$ time to collect $K$ plan update candidates (see the right bottom of Figure~\ref{fig:LP}). 
 
 \textbf{Choosing the best candidate as the plan update:} Next, we verify these candidates on the validation set and select the best one as $\nabla 
 p_t$. Specifically, (1) we first randomly sample a subset of the samples with their label from the training set as the validation set $D_{\textit{valid}}$.
(2) For each plan update candidate, we pseudo-update the task plan by directly adding the plan update candidate into it and test the LLM performance on the validation set with the pseudo-updated plan. If the updated performance is better than the recent average recorded performance by a threshold, we maintain this candidate and its validation performance. Otherwise, we delete it. (3) We choose the candidate with the best validation performance as $\nabla p_t$ and add its performance to the recorded performance list.

\textbf{Compression information:}
However, directly using the generated plan update candidate that consists of multiple new solutions may have the following problems: (1) There may be repeated solutions in the plan update, which should be deleted to reduce the token cost.
 (2) Trival information in the solutions drops the generalization of the plan update toward all samples. 
 (3) The length of the whole task increases with the addition of plan update $\nabla p_t$. Then the total tokens in the input may be beyond the tokens limitation of the LLM.
 To avoid these problems, we prompt the LLM to compress plan update candidates while maintaining its essential information. Specifically, for each task plan update candidate, we use a compression prompt (Figure~\ref{prompt.3}) to prompt the LLM to generate a compressed plan update candidate that only maintains important information.
 Then we choose the best candidate based on the validation performance.
\begin{center}
 \begin{tcbitemize}[raster columns=1,raster equal height,
colframe=black!75!black,colback=black!-15!white,fonttitle=\bfseries]
\tcbitem[squeezed title={Prompt for generating plan update candidates.}]
\fontsize{10}{13}\selectfont{<$m$ Wrong examples>\\
You can generate two or three new correct solutions to avoid the above wrong outputs and to solve all questions refer to the above questions.
\\You must generate solutions different from those previous solutions in the previous natural language plan: $p_{t-1}$.\\
You can generate equations and Python algorithms. \\
When generating one solution, you should write no more than two sentences for one solution. \\
You must not generate detailed examples as we need general solutions}
\end{tcbitemize}
\captionof{figure}{Prompt for obtaining the plan update candidate in the learning to plan method.}
\label{prompt.2}
\end{center} 
 \subsubsection{Updating the Plan}
 \label{sec.update}
 We update the plan by appending the new plan update to the end of the current task plan $p_{t-1}$ (see the right top of Figure~\ref{fig:LP}). If we don't find a new plan update to improve the task plan, we do not update it. 
\begin{center}
 \begin{tcbitemize}[raster columns=1,raster equal height,
colframe=black!75!black,colback=black!-15!white,fonttitle=\bfseries]
\tcbitem[squeezed title={Prompt for compressing the plan update}]
\fontsize{10}{13}\selectfont{You should summarize the similar solutions in [$\nabla p$] into one solution.\\
You should maintain solutions for solving different situations.\\
You must only output no more than five main solutions. \\
When generating one solution, you should write no more than two sentences for one solution.}
\end{tcbitemize}
\captionof{figure}{Compression prompt for the learning to plan method.}
\label{prompt.3}
\end{center}
Too long task plan $p$ has drawbacks: (1) the prompt may be beyond the limitation and (2) turns the short-text inference problem into a long-text inference problem. That increases the task difficulty.
To avoid these, we use $p$, and the compression prompt to prompt the LLM to generate a compressed and shorter $p$. Then we follow the above verification step to identify if we can use the compressed $p$ to replace the original $p$. If the validation performance of the compressed $p$ is better than the original $p$'s performance or just drops a little, we replace the original $p$ with the compressed $p$. If not, we repeat the compression and verification steps until we get a high-quality compressed task plan $p$ or beyond the time limitation.

\textbf{Stop criterion}: We can run our method on the training set with multiple epochs. If we do not find a $\nabla 
 p_t$ to improve the performance during recent batches, we would stop the training process.
\section{Experiment}
\begin{table*}[h]
  \centering
  \scalebox{0.725}{
  \begin{tabular}{c|c|c|c|c| c|c|c|c|c|c|c}
    \hline
    Task &1&2&3&4&5&6&7&8&9&10&Avg
     \\
    \hline ChatGPT (zero-shot, CoT)&73.5&44.0&65.2&30.4&38.8&53.1&29.2&77.0&60.0&12.2&48.3
    \\\hline 
     ChatGPT + Self-Plan (zero-shot, CoT)&83.6&29.6&68.5&30.4&48.3&67.3&45.3&85.5&42.0&15.5&51.6
    \\ \hline
    ChatGPT + Self-Plan with selection (zero-shot, CoT)&86.0&44.3&83.0&30.4&55.1&53.1&73.9&91.7&63.0&17.5&59.8
    \\ \hline
    APE~\cite{zhou2022large}(zero-shot, CoT)
    &76.5&50.2&68.2&35.4&40.1&73.1&39.2&87.2&65.0&15.2&55.0
       \\ \hline
     ChatGPT + Learning-to-Plan~(zero-shot, CoT) &\bf89.8 &\bf55.1&\bf87.0&\bf60.9&\bf56.1&\bf81.0&\bf54.1&\bf96.0&\bf67.0&\bf18.9&\bf66.6\\
    \hline \hline ChatGPT (few-shot, CoT) &79.8&19.6&94.5&43.8&56.1&73.5&75.0&77.1&65.0&56.0&64.0
    \\ 
    \hline
    ChatGPT + Self-Plan (few-shot, CoT)&85.7&29.5&70.3&21.4&52.1&59.2&37.5&97.5&57.0&8.8&51.9
    \\ \hline
    ChatGPT + Self-Plan with selection (few-shot, CoT)&84.0&38.3&97.0&41.4&51.4&65.3&56.5&100.0&75.0&47.3&65.6
    \\ \hline
    APE~\cite{zhou2022large}(few-shot, CoT)
    &82.8&29.3&96.8&50.8&60.1&74.5&76.1&80.1&67.2&60.1&67.8
       \\ \hline
    $k$NN selection~\cite{liu2021makes}~(few-shot, CoT)
    &90.0&31.1&70.6&30.4&55.0&59.2&77.5&87.5&57.0&56.7&61.5
    \\ \hline
    ChatGPT + Learning-to-Plan~(few-shot, CoT)&\bf98.0&\bf38.8&\bf97.8&\bf43.8&\bf56.1&\bf81.6&\bf87.5&\bf100.0&\bf79.0&\bf56.7&\bf73.9\\\hline
  \end{tabular}
  }
  \vspace{2pt}
    \caption{\small{Performance of the 10 tasks on the AMPS  dataset. 'APE' is the Automatic Prompt Engineer method. 'Avg' is the average performance on 10 tasks.}
  }
  \vspace{-10pt}
   \label{tab:Khan}
\end{table*}

\begin{table*}[h]
  \centering
   \scalebox{0.725}{
  \begin{tabular}{c|c|c|c|c|c|c|c|c}
    \hline
    Task &Prealgebra&IA&Algebra&NT&Geometry&Precalculus&CP&Avg
     \\
    \hline ChatGPT (zero-shot, CoT)
    &48.2&13.9&49.0&25.9&17.1&15.4&22.3&27.4
    \\  
    \hline 
   
    ChatGPT + Self-Plan~(zero-shot, CoT) 
    &40.2&11.5&34.2&22.8&17.8&15.5&20.1&23.3
    \\\hline
     ChatGPT + Self-Plan with selection~(zero-shot, CoT) 
    &50.2&14.5&49.3&26.0&18.8&15.5&27.1&28.7
    \\\hline
    APE~\cite{zhou2022large}~(zero-shot, CoT) 
     &49.0&14.9&\bf 50.0&26.8&18.0&16.0&25.3&28.5
    \\\hline
    ChatGPT + Learning-to-Plan~(zero-shot, CoT) 
    &\bf50.6&\bf15.3&49.5&\bf28.2&\bf18.8&\bf16.6&\bf29.3&\bf29.8
    \\\hline \hline
    ChatGPT (few-shot, CoT) &52.4&15.8&49.6&28.3&21.3&16.8&30.2&30.6
    \\ 
    \hline
    ChatGPT + Self-Plan~(few-shot, CoT) 
    &48.9&15.5&48.4&29.3&19.8&\bf16.9&26.2&29.3
    \\\hline
     ChatGPT + Self-Plan with selection~(few-shot, CoT) 
    &53.4&16.0&49.0&29.0&21.5&15.2&29.3&30.5
    \\\hline
        APE~\cite{zhou2022large}(few-shot, CoT)
        &52.4&15.9&\bf50.0&29.0&21.0&16.5&30.0&30.7
         \\\hline
      $k$NN selection~\cite{liu2021makes}~(few-shot, CoT)&
      \bf54.0&14.4&47.5&29.0&20.9&16.0&\bf32.0&30.5\\\hline
    ChatGPT + Learning-to-Plan ~(few-shot, CoT)&53.7&\bf16.9&49.6&\bf29.8&\bf22.5&16.3&30.2&\textbf{31.3}
    \\\hline
  \end{tabular}}
  \vspace{2pt}
 \caption{\small{Performance of the 7 tasks on the Math dataset, 'IA' is the Intermediate Algebra task, and 'NT' is the Number Theory task. 'CP' is the Counting $\&$ Probability task. 'Avg' is the average performance on 7 tasks. }}
 \vspace{-10pt}
  \label{tab:math}
\end{table*}

\begin{table*}[h]
  \centering
   \scalebox{0.705}{
  \begin{tabular}{c|c|c|c|c|c|c|c}
    \hline
    Task &Causal Judgment&CRASS&LSAT&Date Understanding&LLC&SayCan&Avg
     \\
    \hline ChatGPT (zero-shot, CoT)
    &64.5&79.1&52.5&47.3&44.6&36.0&54
    \\  
    \hline 
    ChatGPT + Self-Plan~(zero-shot, CoT)
    &68.2&90.4&52.2&37.3&46.4&36.0&55.1
    \\\hline
     ChatGPT + Self-Plan with selection~(zero-shot, CoT)
    &66.0&91.3&54.0&44.3&50.6&40.0&57.7
    \\\hline
    APE~\cite{zhou2022large}~(zero-shot, CoT) 
   &66.0&83.8&52.9&46.8&60.0&40.0&58.3
    \\\hline
    ChatGPT + Learning-to-Plan~(zero-shot, CoT) 
   &\bf70.8&\bf94.2&\bf55.5&\bf48.0&\bf75.2&\bf52.0&\bf66.0
    \\\hline \hline
    ChatGPT (few-shot, CoT)
    &62.5&88.4&59.2&50.5&72.6&48.0&63.5
    \\ 
    \hline
    ChatGPT + Self-Plan~(few-shot, CoT) 
   &64.5&89.9&43.5&43.0&60.0&44.0&57.5
    \\\hline
     ChatGPT + Self-Plan with selection~(few-shot, CoT) 
   &64.5&91.3&58.5&50.8&75.2&50.0&65.1
    \\\hline
    $k$NN selection~\cite{liu2021makes}~(few-shot, CoT) 
   &58.3&89.6&59.5&\bf53.2&73.0&50.0&63.9
    \\\hline
     APE~\cite{zhou2022large}~(few-shot, CoT) 
   &63.5&91.3&44.9&51.5&75.2&48.0&62.4
    \\\hline
    ChatGPT + Learning-to-Plan~(few-shot, CoT)
    &\bf64.8&\bf92.8&\bf60.1&\bf51.6&\bf81.2&\bf52.0&\bf67.1
    \\\hline
  \end{tabular}}
  \vspace{-2pt}
  \caption{\small{Performance of the non-mathematical reasoning tasks, 'CRASS' is the counterfactual
reasoning assessment task, 'LSAT' is the Logical Reasoning task in LSAT, and 'LLC' is the Last Letter Concatenation task. }}
\vspace{-5pt}
  \label{tab:reason}
\end{table*}

\textbf{Datasets}
{\color{black} \textbf{Mathematical reasoning}, we select 10 challenging mathematical tasks from the AMPS dataset~\cite{hendrycksmath2021} (see Table~\ref{tab:amps_dataset} in Appendix~\ref{appendix.datasets}). Each task corresponds to a specific mathematical problem type and belongs to one of the three fundamental math areas: geometry, calculus, and algebra. 
We also consider the Math dataset~\cite{hendrycksmath2021} and follow its original seven task type annotation: pre-algebra, Intermediate Algebra (IA), Algebra, Counting and Probability (CP), Geometry, Number theory (NT), and Precalculus. \textbf{Causal Reasoning}: We consider the Causal Judgment task from BIG-bench~\cite{srivastava2022beyond}. Given a short story where multiple cause-effect events are introduced, this task asks LLM to answer causal questions such as "Did X cause Y?". We also consider the counterfactual reasoning assessment (CRASS) dataset~\cite{frohberg2021crass}. Given a base premise and a questionized counterfactual
conditional, this task asks LLMs to choose a correct consequence from a set of potential effects. \textbf{Logical Reasoning}: We consider the logical reasoning part of the Law school admission test (LSAT)~\cite{zhong2023agieval}. Given a long passage and a question based on the passage, the LLM needs to answer the question by reasoning. We also consider the date understanding task from BIG-bench~\cite{srivastava2022beyond}, where LLMs need to infer the date based on the date-related information. \textbf{Symbolic reasoning}: We consider the last letter concatenation task~\cite{wei2022chain}, in which LLMs need to extract the last letters of five random sampled words from Wiktionary\footnote{\small\url{https://en.wiktionary.org/wiki/Wiktionary:Frequency_lists/PG/2006/04/1-10000}} and concatenate the last letters as the response.
\textbf{Combinatorial reasoning}: We consider the SayCan dataset~\cite{ahn2022can} where LLMs generate the action sequence for the robot to complete the instruction. More details are in Appendix~\ref{appendix.datasets}.}

\textbf{Model and settings}: We use the ChatGPT model (GPT-3.5-turbo) as the LLM as the trade-off of performance and inference cost \footnote{ \small Link:~\url{https://platform.openai.com/}}. We consider the zero-shot CoT and few-shot CoT settings. In the zero-shot CoT test setting, we follow~\cite{NEURIPS2022_8bb0d291} and add the chain of thought (CoT) prompt 'Let's think step by step' after the test sample. In the few-shot CoT test setting, we use the few-shot CoT prompt that contains fixed demonstration examples and the test problem.  Each example consists of the example problem, the explanation (solution) for this problem, and the correct answer. We follow~\cite{lewkowycz2022solving,wei2022chain} and set four examples for tasks of mathematical reasoning, logical reasoning, symbolic reasoning, and causal reasoning and six examples for the combinatorial reasoning task. 
~The examples are randomly chosen from the original training set.
For the AMPS and Math datasets, we use specific prompts for each category for all methods. Our method uses the same examples in both training and test settings. For our method and the self-plan method, we further add the learned plan to the prompt. We list the zero-shot/few-shot inference prompt we used in Appendix~\ref{appendix.zero-few}.

\textbf{Baselines}
We consider these baselines (1) the zero-shot/few-shot performance of the LLM directly measured in the test set. (2) The self-plan method: We use the task plan generated by the self-plan method to guide the inference of the LLM and record its zero-shot/few-shot test performance. (3) The self-plan method with selection: We generate multiple plan candidates by the self-plan method and choose the one with the best performance as the task plan. (4)~$k$NN selection method~\cite{liu2021makes}: For every test example, we use the BM25 algorithm to retrieve $k$ most similar examples from the training set as the demonstration examples in the prompt. This method utilizes the entire training set like our method. (5)~Automatic Prompt Engineer~\cite{liu2021gpt} that iteratively updates the task instruction by generating instruction candidates and then selecting the one with the best validation performance. We tried using the self-refine method~\cite{madaan2023self} as a baseline, but its templates are tailored for specific tasks and don't have a suitable template for our non-mathematical tasks. For math tasks, its GSM8k~\cite{cobbe2021training} template is not suitable for non-arithmetic math tasks. So we just discuss it. 

\begin{table*}[htp]
  \centering
     \scalebox{0.705}{\begin{tabular}{c|c|c|c|c|c}
    \hline
    Task &Algebra &Causal Judgment&Logical Reasoning (LSAT)&Last letter concatenation &SayCan
     \\
    \hline GPT-4-32k
    &57.8&64.6&81.3&85.4&72.0
    \\  
    \hline 
    GPT-4-32k + Learned-Plan $p$ from ChatGPT
    &\bf58.9&\bf68.8&\bf82.0&\bf88.6&\bf80.0
    \\\hline
  \end{tabular}}
  \vspace{2pt}
  \caption{Five different reasoning tasks' zero-shot CoT performance. 'Algebra' task is from the math dataset.}
  \vspace{-10pt}
  \label{tab:transfer}
\end{table*}
\begin{table*}[h]
  \centering
  \scalebox{0.685}{\begin{tabular}{c|c|c|c|c|c}
    \hline
    Task &Composite function&CRASS&Logical Reasoning&Last letter concatenation&SayCan
     \\
    \hline Vicuna (zero-shot, CoT)
    &10.2&46.4&31.0&2.0&22.0
    \\  
    \hline 
    Vicuna + Learned-Plan $p$ from Vicuna~(zero-shot, CoT)
    &16.1&94.3&37.8&3.2&25.2
    \\\hline
    \hline WizardLM (zero-shot, CoT)
    &12.3&87.0&18.3&6.8&25.2
    \\  
    \hline 
    WizardLM + Learned-Plan $p$ from Vicuna~(zero-shot, CoT)
    &15.0&98.5&19.4&13.8&30.2
    \\\hline
  \end{tabular}}
  \vspace{2pt}
  \caption{\small{Performance of five different type reasoning tasks. 'CRASS' is the Counterfactual Reasoning Assessment task. 'Composite function' is the second task from the AMPS dataset. 'Logical Reasoning' is from the Law school admission test. }}
  \vspace{-10pt}
  \label{tab:open}
\end{table*}
\begin{table*}[h]
  \centering
   \scalebox{0.63}{\begin{tabular}{c|cccc|ccc|ccc|cc}
    \hline
    \multicolumn{1}{c|}{Hyper-parameter}&\multicolumn{4}{c|}{CoT}&\multicolumn{3}{c|}{Epoch number}&\multicolumn{3}{c|}{Candidate number $K$}&\multicolumn{2}{c}{Candidate compression}\\
    \hline
    {Value}&{w/o}&{w/o+our method}&{CoT}&{CoT+our method}&{1}&{5}&{10}&{1}&{5}&{10}&{w/o}&{our method}
     \\
    \hline
    Composite function~(AMPS dataset) &42.8&51.0&44.0&55.1&47.1&50.1&55.1&50.0&55.1&55.1&42.8&55.1
    \\  
    \hline 
    Causal Judgment&56.3&60.4&64.5&70.5&64.8&68.7&70.5&69.0&70.5&70.5&58.1&70.5
    \\\hline
    Last letter concatenation&40.2&60.2&44.6&75.2&58.7&70.0&75.2&52.8&75.2&77.8&50.5&75.2
    \\\hline 
  \end{tabular}}
  \vspace{2pt}
  \caption{\small{Ablation study.}}
  \vspace{-10pt}
  \label{tab:ablation1}
\end{table*}

\begin{table*}[h!]
  \centering
   \scalebox{0.7}{\begin{tabular}{c|ccc|ccc|ccc|cccc}
    \hline
    \multicolumn{1}{c|}{Hyper-parameter}&\multicolumn{3}{c|}{Validation size}&\multicolumn{3}{c|}{Wrong samples $m$}&\multicolumn{3}{c|}{Threshold}&\multicolumn{4}{c}{Plan compression frequency}\\
  
    \hline
    {Value}&{1}&{5}&{10}&{1}&{5}&{10}&{1}&{5}&{10}&{1}&{3}&{5}&{w/o}
     \\
    \hline
    Composite function~(AMPS dataset) &41.8&55.1&57.0&36.7&55.1&47.1&55.1&44.0&44.0&53.0&55.1&47.1&45.2\\  
    \hline 
    Causal Judgment&54.2&70.5&71.0&58.3&70.5&64.8&70.5&66.7&64.5&70.5&70.5&62.0&55.0 
    \\\hline
    Last letter concatenation&62.4&75.2&75.2&70.6&75.2&54.2&75.2&63.2&60.6&76.6&75.2&70.2&58.7
    \\\hline 
  \end{tabular}}
  \vspace{2pt}
  \caption{\small{Ablation study. The validation size is equal to the time value multiplied training batch size.}}
  \vspace{-10pt}
  \label{tab:ablation2}
\end{table*}

\textbf{Hyper-parameters}
For our method, we set the training batch size as 32 and run each task with 10 epochs. We stop the training process if the task plan is not updated in the recent 10 batches. We set the validation set size to be 5 times that of the training batch size. We set the number of selected wrong samples $m$ and the time of generating revision candidates $K$ in Section~\ref{sec.revision} as 3 and 5, respectively. We set the threshold as 1.0 and the compression frequency of the plan as 3. We use a temperature of 0 for stable output. For simple tasks (e.g., the AMPS tasks), we ask LLMs to summarize the plan into no more than five main general solutions, and for complex tasks (e.g., Math dataset), the number is ten. The recent average recorded performance in Section~\ref{sec.revision} is the average of the recent three recorded validation performances.  Ablation studies of these hyper-parameters are in Sec.~\ref{sec.ablation}.
We put the experiment details of baselines in Appendix~\ref{appendix.baselines}.
\subsection{Experiment Results}
{\color{black}Zero-shot CoT setting: {\color{black}We observe that our method obviously improves the LLM's average zero-shot performance of 10 mathematical AMPS tasks by 18.3 percent (Table~\ref{tab:Khan}) and achieves the highest performance (29.5) for the complex math tasks from the Math dataset (Table~\ref{tab:math}). Our method also improves the performance of four non-mathematical reasoning tasks markedly (Table~\ref{tab:reason}), especially for the last letter concatenation task (from 44.6 percent to 75.2 percent). The huge improvement over reasoning tasks is because LLMs learn a high-quality task plan from the training set to solve complex tasks.} In contrast, the self-plan method's generated plan shows significant performance instability. For example, this method drops the average performance on the math dataset by 4.1 percent. The generated plan contains many unverified factual errors, rendering it impractical. Selecting the task plan based on the validation performance to reduce the factual error boosts the method's performance from 23.3 to 28.7\% on the math dataset. Our method still outperforms the self-plan method with selection as our method further has the advantage of iteratively learning from the error that LLMs can not solve. Our method outperforms APE as it only optimizes the task instruction and does not learn from errors.

Few-shot CoT setting: Experiments in Table~\ref{tab:Khan}, ~\ref{tab:math}, and ~\ref{tab:reason} shows that our method can further improve few-shot performance by learning a task plan to solve the errors that have not been solved with the guidance of demonstration examples. Our method's average performance also outperforms other baselines by a clear margin. The performance improvement of the self-plan method is still unstable as the factual error problem and can be improved by the selection strategy. APE influences the few-shot performance slightly as it only adjusts the task instruction. $k$NN selection method improves the LLM's few-shot performance slightly as the most similar examples based on the term frequency may not mean the best demonstration examples for complex reasoning tasks. } 
\subsection{Analysis}
\vspace{-2pt}
\textbf{Transferring from one LLM to another LLM}
Humans can directly transfer knowledge through natural language. Surprisingly, we find that the learned task plan from the ChatGPT model can improve the zero-shot performance of GPT-4 in Table~\ref{tab:transfer}. That means when both LLMs can understand natural language well, the learned task plan can be directly transferred from LLM A to LLM B to improve LLM B's performance as natural language task knowledge (step-by-step solutions) rather than a simple task prompt. 

\textbf{Improving the open-source LLM's performance}
{\color{black} We deploy our method with the open-source vicuna-13b-v1.5 model~\cite{chiang2023vicuna} (based on Llama-2~\cite{touvron2023llama}) to learn the task plan of main reasoning tasks. Table~\ref{tab:open} shows that the learned plan by the newest Vicuna model can boost its zero-shot test performance, which means our method also works well for strong open-source LLMs. The learned plans by the Vicuna model are clear but shorter and simpler than those learned by ChatGPT in Appendix~\ref{appendix.vicuna}.}  

\textbf{Quality analysis of the learned task plan and the continual update process}
We present the learned task plans by ChatGPT in Appendix~\ref{appendix.gpt_4}. We find that the general solutions in the learned plan are highly relevant and detailed for solving task samples and are diverse for solving different cases. Figure~\ref{analysis:continual} in Appendix~\ref{appendix.gpt_4} shows that our method learns two general solutions in the plan during the training. However, the plan is not yet complete and some test samples can not be solved. After the training process, the trained plan has rich general step-by-step solutions to calculate angles and achieve success in solving the test samples. That shows our method's ability to improve the plan quality by learning from errors. 

\textbf{Comparing our learning method's plan with that of the self-plan method} 
We compare the learned plan with that of the self-plan method 
in Figure~\ref{fig: plans} in Appendix~\ref{appendix.case}.  The self-plan's generated task plan has a factual error about fraction simplification. The self-plan with selection method provides a plan that contains correct solutions related to the greatest common factor. Compared to them, our learned task plan provides verified and diverse step-by-step solutions for different situations, such as mixed numbers and fractions with variables. This is because our plan learns from various samples and errors that LLM may encounter. 

\subsection{Ablation Study}
\label{sec.ablation}
We conduct the ablation experiments for various factors (1)~\textbf{CoT prompt}: The performance of experiments without the CoT prompt drops slightly (see Table~\ref{tab:ablation1}) and our method can still improve LLM's performance (see experiment 'w/o+our method') as our method improves the performance by learning from LLM's. We believe our method can improve other prompting methods by learning from errors. (2)~\textbf{Epoch number}: we record the performance in the 1st, 5th, and 10th epoch. From Table~\ref{tab:ablation1}, we observe that the performance arises quickly at first epochs and then rises slowly. So we set the number of epochs as 10. (3)~\textbf{Candidate number $K$ (Sec~\ref{sec.revision})}: We find a positive correlation between $K$ and the performance as more candidates mean a larger search space to find a stronger plan update. But more candidates also means higher computational consumption required to verify them. To trade off two aspects, we set $K$ as 5. (4)~\textbf{Candidate compression} (Sec~\ref{sec.revision}): Compressing the candidate indeed boosts the performance (see the last chunk in Table~\ref{tab:ablation1}) as it makes the plan update candidate more general and becomes applicable for all task samples.
(5)~\textbf{Validation size}: A larger validation size results in better generalization performance (see Table~\ref{tab:ablation2}) with a higher computation cost inference cost. Therefore, we set the validation size as 5 times the training batch size to trade off two aspects. 
(6)~\textbf{$m$ wrong samples} in the plan update generation prompt (Figure~\ref{prompt.2}): In Table~\ref{tab:ablation2}, we observe that performance first improves with larger $m$ as LLMs can refer to more wrong examples and then it drops as the poor induction ability of the ChatGPT for too many examples. (7)~\textbf{threshold} in selecting the best candidate (Sec~\ref{sec.revision}): We set the threshold as 1.0 as a higher threshold may filter out good revision, as shown in Table~\ref{tab:ablation2}. (8) \textbf{Plan compression frequency} in (Sec.~\ref{sec.update}): A higher compression frequency enhances performance, as shown in Table~\ref{tab:ablation2} under ' plan compression frequency' due to a higher quality plan and reduced memory requirements. We set the compression frequency to 3 to trade off the performance and the compression inference cost. 
\section{Conclusion}
Large Language Models (LLMs) have shown impressive performance in various natural language tasks. But they still need a task plan when processing complex tasks. In this paper, we introduce the learning to plan method, which explicitly learns the task plan based on the error feedback from the training dataset and uses the learned task plan to solve the task. Our proposed method's effectiveness has been verified through multiple reasoning tasks, and we conclude that it is a promising approach for improving LLMs' complex task performance. 
\section{Limitations and potential risks}
\label{appendix.limit}
Although our learning-to-plan method achieves significant improvement on multiple reasoning tasks. It still has some limitations:
(1) The current learning-to-plan method assumes that the test samples are derived from well-defined tasks, making it challenging to deploy in the open-domain setting. 
In the future, we will investigate this problem.
(2) Errors such likes misunderstanding specific words, terms, and numbers are hard to solve by our method. Continually updating LLMs' parameters to improve their basic perception ability is still necessary. 
(3) A theoretical guarantee for the convergence of our method is needed. We will study it in the future.

We do not see any potential risk as our experiments are conducted on public datasets and our method focuses on reasoning tasks.
\bibliography{anthology,custom}

\appendix
\section{Prompt for self-plan method}
\label{appendx.self-plan}
We illustrate the prompt for the self-plan method in Figure~\ref{prompt.1}. For each task, we use the task description from Table~\ref{tab:amps_dataset},Table~\ref{tab:mathdataset}, and Table~\ref{tab:reasondataset}.
\begin{center}
\begin{tcbitemize}[raster columns=1,raster equal height,
colframe=black!75!black,colback=black!-15!white,fonttitle=\bfseries]
\tcbitem[squeezed title={Self-plan method's prompt}]
<Task description>\\
You can generate multiple general solutions to solve any questions from the above task.\\You can consider equations and algorithms. \\When generating one solution, you should write no more than two sentences for one solution. \\Solutions:
\end{tcbitemize}
\captionof{figure}{Prompt for the self-plan method.}
\label{prompt.1}
\end{center}
\begin{table*}[htp]
  \centering
  \caption{\small{The summary of the 10 tasks on the AMPS  dataset. 'N' is the number of examples in one task.}}
  \tiny
  \begin{tabular}{cccp{4cm}p{5cm}}
    \hline
    Task id &N&Task type&Task description&Example
     \\\hline
     1&199&Calculus&Fraction division in simplest form &Reduce to lowest terms:$\frac{1}{9}\div\frac{7}{6}$
      \\\hline
        2&394&Calculus&Calculating the value of a composite function&$h(n) = 2n$, $f(x) = 2x^{2}+6x-h(x)$, $ f(h(-2)) = {?} $
      \\\hline
      3&368&Geometry&Angle Calculation& If $\angle AOC = 180^\circ$ and $\angle BOC = 160^\circ$, what is $\angle AOB$, in degrees? 
      \\\hline
      4&94&Geometry&Calculating the length of the hypotenuse or a leg of the right triangle&In the right triangle shown,  angle $A=30^\circ$ and $AC=18$. How long is $AB$? 
      \\\hline
      5&393&Algebra&Given a number A, find its factor/multiple in a list of numbers. &Which of the following numbers is a factor of 56? $\{3,6,8,10,13\}$
      \\\hline
      6 &197&Calculus&Calculating the significant figures in a number&How many significant figures does $00.033$ have? 
      \\\hline
      7&97&Geometry&Calculating the sin/cos value of an angle based on the length of legs and other conditions.&$\overline{AC}$ is $8$ units long, $\overline{BC}$ is $15$ units long, $\overline{AB}$ is $17$ units long, What is $\cos(\angle ABC)$? 
in a triangle. 
      \\\\\hline
      8&192&Calculus&Fraction Division&$\frac{1}{8} \div \frac{8}{3} = {?}$ 
      \\\hline
      9&400&Calculus&Fraction reduction in the form of the mixed number&Express your answer as a mixed number simplified to lowest terms. $9\dfrac{3}{4}-5\dfrac{4}{6} = {?}$
      \\\hline
      10&593&Geometry&Calculating the length of a leg in a triangle based on the length of another leg and the sin, cos, and tan value of an angle& $\overline{BC}=9$, $\sin(\angle ABC )=\frac{ \sqrt{2}}{2}$,$\cos( \angle ABC )=\frac{\sqrt{2}}{2}$, $\tan( \angle ABC )=1$, then $\overline{AC}={?}$
      \\\hline   
  \end{tabular}
  \vspace{-5pt}
  \label{tab:amps_dataset}
\end{table*}
\begin{table*}[htp]
  \centering
   \caption{\small{The summary of the 7 tasks on the Math  dataset. 'Training number' is the number of examples in the training set of one task. 'Test number' is the number of examples in the test set of one task.}}
  \tiny
  \begin{tabular}{cccp{3cm}p{4.5cm}}
    \hline
    Task name &Training number&Test number&Task description&Example
     \\\hline
     Prealgebera&1205&871&basic-level algebra problem&What is the largest prime factor of 78?
      \\\hline
        Intermediate Algebra&1295&903&intermediate-level algebra problem&Let $a,$ $b,$ and $c$ be distinct real numbers such that$\frac{a^3 + 6}{a} = \frac{b^3 + 6}{b} = \frac{c^3 + 6}{c}$. Find $a^3 + b^3 + c^3$.
      \\\hline
      Algebra&1187&1744&high-level algebra problem&For what value of $x$ does $2^{12} = \left(\frac{1}{8}\right)^x$?
      \\\hline
      Number Theory&869&540&Number Theory problem&What is the sum of the four positive factors of the positive integer value of $\sqrt{196}$?
      \\\hline
      Counting and Probability &771&474&probability problem&A box contains 5 white balls and 6 black balls.  Five balls are drawn out of the box at random.  What is the probability that they all are white?
      \\\hline
      Geometry&870&479&Geometry problem&What is the sum of the lengths of the \text{altitudes} of a triangle whose side lengths are $10,$ $10,$ and $12$? Express your answer as a decimal to the nearest tenth.
      \\\hline
      Precalculus&746&546&basic level calculus problem&Find all values of $k$ for which the angle between the vectors  $\begin{pmatrix} $k$ \\ 1 \\ 1 \end{pmatrix}$ and $\begin{pmatrix} 1 \\ k \\ 1 \end{pmatrix}$  is $\frac{\pi}{3}$.
      \\\hline
     
  \end{tabular}
  \vspace{-5pt}
  \label{tab:mathdataset}
\end{table*}
\begin{table*}[htp]
  \centering
   \caption{\small{The summary of the 6 reasoning tasks. 'Training number' is the number of examples in the training set of one task. 'Test number' is the number of examples in the test set of one task.}}
  \tiny
  \begin{tabular}{cccp{3cm}p{4.5cm}}
    \hline
    Task name &Training number&Test number&Task description&Example
     \\\hline
     Causal Judgment &189&47&Given a short story, LLMs need to judge if event X causes event Y. &Story: ... Question: Did Bob intentionally harm the health of the townspeople?
      \\\hline
     Counterfactual Reasoning Assessment&273&68&Given a premise, LLMs need to judge the possible result if the event in the premise does not happen.&Premise: A man does not flirt with a woman. Question: What would have happened if he had flirted with her? Options: (A) The man would have asked the woman on a date. (B) The man would not have asked the woman on a date. (C) That is not possible.
      \\\hline
      Logical Reasoning (LSAT)&3503&509&Given a passage and a question, LLMs need to infer which option is true. & Passage: ... Question: Which one of the following principles is best illustrated by the example above? Options: (A) ... (B) ... (C) ... (D) ...
      \\\hline
      Date Understanding &368&92& LLMs need to infer the date based on the context. &The first day of 2019 is a Tuesday, and today is the first Monday of 2019. What is the date 24 hours later in MM/DD/YYYY?
      \\\hline
      Last Letter Concatenation&1500&500&Given n words, LLMs need to extract and concatenate the last letters of the n words. &Take the last letters of each word in "California endurance drink finely singing" and concatenate them
      \\\hline
      SayCan&73&25&Given an instruction, LLMs need to generate an action sequence for the robot to complete the instruction.& How would you move the chips bag from the table to the counter?
      \\\hline
  \end{tabular}
  \vspace{-5pt}
  \label{tab:reasondataset}
\end{table*}
\section{Dataset Details}
\label{appendix.datasets}
For the AMPS pre-training dataset~\cite{hendrycksmath2021} (see Table~\ref{tab:amps_dataset}). We exclude tasks that have only a few data points (e.g., less than 50) and select tasks that are not fully solved by the ChatGPT model. For tasks in the AMPS pre-training dataset, the Causal Judgement task, and the date understanding task, we split the entire dataset into the training set and the test set using a 3:1 ratio. When using the learning-to-plan method to verify the revision candidates, we randomly sample some labeled samples from the training set as the validation set. For other datasets (Table~\ref{tab:mathdataset} and Table~\ref{tab:reasondataset}), we construct the training set using its respective training data from the original training set. We use the test data from the original test set to construct the test set. 

\section{The zero-shot and few-shot inference prompts}
\label{appendix.zero-few}
We list the zero-shot and few-shot inference prompts in Figure~\ref{prompt.zero} and~\ref{prompt.few} and provide an example for them respectively. The task plan in the prompt guides the LLM to solve the complex task step by step. For baselines that do not have a plan, we directly use the inference prompts without the plan. 
\begin{figure*}
 \begin{tcbitemize}[raster columns=1,raster equal height,
colframe=black!75!black,colback=black!-15!white,fonttitle=\bfseries]
\tcbitem[squeezed title={Our method's zero-shot inference prompt for the test sample.}]
\textbf{<Prompt pattern>}\\
$[$Question$]$ 

Let's think step by step. $[$Plan$]$. Please follow the solutions step by step to get the answer. 
\\
\textbf{<Example>}\\
"Passage: Retailers that excel in neither convenience nor variety of merchandise tend not to be very successful. Yet many successful retailers excel in just one of the areas and meet competitors' standards for the other. Hence, a retailer's success need not depend on excellence in both areas. Question: The structure of the reasoning in the argument above is most parallel to that in which one of the following? Options: (A)Runners who have only average speed and endurance are unlikely to win long-distance races. Some long-distance champions, however, win by being above average in speed or endurance only; therefore, being above average in both speed and endurance is not necessary. (B)Bicyclists who have only average speed are unlikely to win short races, but in a long-distance race such bicyclists can win if they have better-built bicycles than average and better endurance than average. Therefore, most bicycle races are not won by bicyclists with above-average speed. (C)Excellence in a particular swimming stroke is not always necessary in order for a swimmer to win a race that requires each swimmer to use several different strokes in sequence, and many swimmers win these races without being the best at any of the strokes. Therefore, anyone who does excel at all the strokes is almost certain to win. (D)Apples that are neither especially firm nor especially flavorful are unsuitable for baking; yet while flavor is essential for both baking and eating, many flavorful apples that are soft are suitable for eating. Hence, the apples that are best for eating need not be both firm and flavorful. (E)Most plants that are neither ornamental nor edible are useless and are thus classified as weeds; yet many such plants are useful for purposes other than food or ornamentation, and are thus not classified as weeds. Hence, not all inedible and non-ornamental plants are weeds. 

Let's think step by step. You need to identify the conclusion and premises of an argument and look for any assumptions that the argument relies on. Consider any counterexamples or objections that could weaken the argument. Look for evidence or reasoning that supports or undermines the conclusion. Identify the conclusion and premises of the argument, and look for any assumptions it relies on. Consider any counterexamples or objections that could weaken the argument. Look for evidence or reasoning that supports or undermines the conclusion. Please follow the solutions step by step to get the answer."

\end{tcbitemize}
\captionof{figure}{Zero-shot inference prompt. We give the general prompt pattern we used for all tasks first and then we give an example from the logical reasoning task to explain the general prompt pattern. We put the learned plan into '[plan]'.}
\label{prompt.zero}
\end{figure*}
\FloatBarrier
 \begin{figure*}[h]
 \begin{tcbitemize}[raster columns=1,raster equal height,
colframe=black!75!black,colback=black!-15!white,fonttitle=\bfseries]
\tcbitem[squeezed title={Our method's few-shot prompt for the test sample.}]
\textbf{<Prompt pattern>}
Here are the answers to the problems in the exam.

Problem 1: [Problem description]

Explanation for Problem 1: [Problem explanation]

The answer is therefore: [Answer]

...

Problem n: [Problem description]

Explanation for Problem n: [Problem explanation]

The answer is therefore: [Answer]

Problem n+1: [Problem description]

Let's think step by step. [Plan]. Please follow the solutions step by step to get the answer.

Explanation for Problem n+1:
\end{tcbitemize}
\captionof{figure}{Few-shot inference Prompt. We give the general few-shot prompt pattern we used for all tasks first and n is the number of demonstration examples.}
\label{prompt.few}
\end{figure*} \FloatBarrier
 \begin{figure*}[h] 
\centering
 \begin{tcbitemize}[raster columns=1,raster equal height,
colframe=black!75!black,colback=black!-15!white,fonttitle=\bfseries]
\tcbitem[squeezed title={Example of our method's few-shot prompt.}]
\textbf{<Example>}

Here are the answers to the problems in the exam.

Problem 1:$h(n) = 4n^{2}-2(f(n))$ $f(n) = n+1$ $g(x) = 5x-4(f(x))$ $ f(h(0)) = {?} 
$

Explanation for Problem 1:First, let's solve for the value of the inner function,$h(0)$. Then we'll know what to plug into the outer function.$h(0) = 4(0^{2})-2(f(0))$To solve for the value of$h$, we need to solve for the value of$f(0)$$f(0) = 1$$f(0) = 1$That means$h(0) = 4(0^{2})+(-2)(1)$$h(0) = -2$Now we know that$h(0) = -2$. Let's solve for$f(h(0))$, which is$f(-2)$$f(-2) = -2+1$$f(-2) = -1$ 

The answer is therefore -1. 

Problem 2:$g(t) = 6t-f(t)$ $f(t) = 2t$ $ g(f(9)) = {?} 
$

Explanation for Problem 2:First, let's solve for the value of the inner function,$f(9)$. Then we'll know what to plug into the outer function.$f(9) = (2)(9)$$f(9) = 18$Now we know that$f(9) = 18$. Let's solve for$g(f(9))$, which is$g(18)$$g(18) = (6)(18)-f(18)$To solve for the value of$g$, we need to solve for the value of$f(18)$$f(18) = (2)(18)$$f(18) = 36$That means$g(18) = (6)(18)-36$$g(18) = 72$

The answer is therefore 72. 

Problem 3:$g(x) = -3x^{2}-2x-7+4(f(x))$ $f(x) = -6x^{2}-2(h(x))$ $h(x) = -4x^{2}+3x$ $ h(f(2)) = {?} $

Explanation for Problem 3:First, let's solve for the value of the inner function,$f(2)$. Then we'll know what to plug into the outer function.$f(2) = -6(2^{2})-2(h(2))$To solve for the value of$f$, we need to solve for the value of$h(2)$$h(2) = -4(2^{2})+(3)(2)$$h(2) = -10$That means$f(2) = -6(2^{2})+(-2)(-10)$$f(2) = -4$Now we know that$f(2) = -4$. Let's solve for$h(f(2))$, which is$h(-4)$$h(-4) = -4(-4)^{2}+(3)(-4)$$h(-4) = -76$

The answer is therefore -76. 

Problem 4:$h(t) = -4t^{2}+4t-2-2(g(t))$ $g(x) = 7x$ $ g(h(1)) = {?} $

Explanation for Problem 4:First, let's solve for the value of the inner function,$h(1)$. Then we'll know what to plug into the outer function.$h(1) = -4(1^{2})+(4)(1)-2-2(g(1))$To solve for the value of$h$, we need to solve for the value of$g(1)$$g(1) = (7)(1)$$g(1) = 7$That means$h(1) = -4(1^{2})+(4)(1)-2+(-2)(7)$$h(1) = -16$Now we know that$h(1) = -16$. Let's solve for$g(h(1))$, which is$g(-16)$$g(-16) = (7)(-16)$$g(-16) = -112$

The answer is therefore -112. 

Problem 5:$h(n) = -6n-1$ $g(x) = 7x^{2}+7x+7-5(h(x))$ $ g(h(1)) = {?} $

Let's think step by step. When substituting a value into a function that is defined in terms of another function, make sure to substitute the entire expression for the second function, not just the value of the variable. Please follow these solutions step by step to get the answer.

Explanation for Problem 5:

\end{tcbitemize}
\captionof{figure}{Example for the few-shot inference Prompt from the math reasoning task. n is the number of demonstration examples.}
\label{prompt.few_example}
\end{figure*} 
\FloatBarrier

\section{Details of Baselines}
For all baselines, we use the same ChatGPT model with our learning to plan method. For the self-plan with selection method, the candidate number is equal to $K\times$ \textit{total training iteration number in the learning to plan method} for a fair comparison. The size of the validation set is equal to that of our method. For other baselines, we follow their official code to conduct experiments across all datasets\footnote{ APE: \url{https://github.com/keirp/automatic_prompt_engineer}}. For the self-plan with selection method, the candidate number is equal to $K\times$ \textit{total training iteration number in the learning to plan method} for a fair comparison.
\label{appendix.baselines}
\FloatBarrier
\section{The learned task plan by Vicuna}
\label{appendix.vicuna}

 \begin{center}
\begin{tcolorbox}[enhanced,title=Task 2,
attach boxed title to top left=
{xshift=2mm,yshift=-2mm},
boxed title style={size=small,colback=black},colback=black!-15!white]
\textbf{Task Definition}: Calculating the value of a composite function (dataset size:394).\\\\\textbf{Example}: $h(n) = 2n$, $f(x) = 2x^{2}+6x-h(x)$, $ f(h(-2)) = {?} $\\\\
\textbf{Learned natural language plan $p$ (zero-shot cot)}:
\\\\\small{Here are the main solutions for solving the given problem:           
\\\\                                  
1. Simplify the given functions and substitute the given values into the equation to calculate the required value.\\\\                         
2. Differentiate the given functions if necessary, substitute the given values into the function, and calculate the required value.     \\\\                   
3. Use the quadratic formula to solve the quadratic equation.
\\\\
4. Determine the type of equation and use the appropriate method to solve for the required value.                                          \\\\             
5. Factor the equation if necessary and solve for the required value. } 
\\\\
\end{tcolorbox}
\captionof{figure}{The task details of the 'composite function' task of the AMPS dataset and its learned task plan learned by Vicuna}
\label{program:task2_vicuna}
\end{center} 
\begin{figure}
\begin{tcolorbox}[enhanced,title=Counterfactual Reasoning Assessment Task,
attach boxed title to top left=
{xshift=2mm,yshift=-2mm},
boxed title style={size=small,colback=black},colback=black!-15!white]
\textbf{Task Definition}: Given a premise, LLMs need to
judge the possible result if the event
in the premise does not happen. \\\\\textbf{Example}: Premise: A woman does not increase his own capacity. 

Question: What would have happened if she had increased his own capacity?

Options: (A) The woman would have felt accomplished. (B) The woman would not have felt accomplished. (C) That is not possible.
\\\\
\textbf{Learned natural language plan $p$ (zero-shot cot)}:
\\\\\small{
When dealing with anti-causal reasoning questions, it is important to consider all possible outcomes and eliminate any options that are not supported by the premise. In some cases, it may be necessary to consider additional information or assumptions to make an informed decision.
\\\\
When dealing with anti-causal reasoning questions, it is important to carefully analyze the premise and consider all possible scenarios before selecting an answer. It is also important to avoid making assumptions or jumping to conclusions based on common sense or intuition, as these may not always be accurate. Instead, it is important to consider all possible outcomes and eliminate any options that are not supported by the premise. In some cases, it may be necessary to consider additional information or assumptions to make an informed decision.
} 
\end{tcolorbox}
\captionof{figure}{The task details of the Counterfactual Reasoning Assessment task and its task plan learned by Vicuna.}
\label{program:crass_vicuna}
\end{figure} 
 \begin{figure*}[h]
\begin{tcolorbox}[enhanced,title=Last Letter Concatenation Task,
attach boxed title to top left=
{xshift=2mm,yshift=-2mm},
boxed title style={size=small,colback=black},colback=black!-15!white]
\textbf{Task Definition}: Given n words, LLMs need to extract and concatenate the last letters
of the n words. \\\\\textbf{Example}: Take the last letters of each words in "spur drowning Japan dialect valet " and concatenate them \\\\
\textbf{Learned natural language plan $p$ (zero-shot cot)}:
\\\\\small{
To concatenate the last letters of each word in a phrase, follow these steps: identify the last letter of each word, arrange the letters in the correct order, and concatenate them to form a new word. To avoid errors, it is important to pay attention to the position of the words in the phrase and double-check your work.
} 
\end{tcolorbox}
\captionof{figure}{The task details of the Last Letter Concatenation task and its task plan learned by Vicuna}
\label{program:last_letter_vicuna}
\end{figure*} \FloatBarrier
 \begin{figure*}[h]
\begin{tcolorbox}[enhanced,title=Logical Reasoning (LSAT) Task,
attach boxed title to top left=
{xshift=2mm,yshift=-2mm},
boxed title style={size=small,colback=black},colback=black!-15!white]
\textbf{Task Definition}: Given a passage and a question,
LLMs need to infer which option
is true. \\\\\textbf{Example}: Passage: Hana said she was not going to invite her brothers to her birthday party. However, among the gifts Hana received at her party was a recording in which she had expressed an interest. Since her brothers had planned to give her that recording, at least some of Hana's brothers must have been among the guests at Hana's birthday party after all.

Question: A reasoning error in the argument is that the argument

Options: (A) disregards the possibility that a change of mind might be justified by a change in circumstances (B)treats the fact of someone's presence at a given event as a guarantee that that person had a legitimate reason to be at that event (C)uses a term that is intrinsically evaluative as though that term was purely descriptive (D)fails to establish that something true of some people is true of only those people (E)overlooks the possibility that a person's interest in one kind of thing is compatible with that person's interest in a different kind of thing.
\\\\
\textbf{Learned natural language plan $p$ (zero-shot cot)}:
\\\\\small{
To avoid errors in logical reasoning, carefully examine the premises and conclusions of an argument and identify any flaws or unsupported assumptions. Consider alternative explanations and potential counterarguments to strengthen the evaluation of the argument.
\\\\
To solve logical reasoning questions, carefully analyze the premises and conclusions of the argument, look for any flaws or unsupported assumptions, and consider alternative explanations and potential counterarguments. Evaluate each option based on how well it supports the conclusion, and avoid errors by maintaining separate solutions for different situations. When generating a solution, write no more than two sentences.
} 
\end{tcolorbox}
\captionof{figure}{The task details of the Logical Reasoning (LSAT) task and its task plan learned by Vicuna}
\label{program:lsat_vicuna}
\end{figure*} \FloatBarrier

 \begin{figure*}[h]
\begin{tcolorbox}[enhanced,title=SayCan Task,
attach boxed title to top left=
{xshift=2mm,yshift=-2mm},
boxed title style={size=small,colback=black},colback=black!-15!white]
\textbf{Task Definition}: Given an instruction, LLMs need to
generate an action sequence for the
robot to complete the instruction. \\\\\textbf{Example}: How would you move the chips bag from the table to
the counter?
\\\\
\textbf{Learned natural language plan $p$ (zero-shot cot)}:
\\\\\small{
solutions: To locate a missing item, we can follow these general rules:
\\\\
1. Check all possible locations for the item and bring it to the user if found.
\\\\
2. Use a search algorithm or consult with others to find the item and bring it to the user.
\\\\
3. Ask the user to describe the item in more detail to narrow down the search.
\\\\
4. Review the item's usage history or consult with other users to determine its typical location.
\\\\
5. Use a tracking system or designated storage location to locate the item.
\\\\
6. Consider the urgency of the situation and prioritize the search accordingly.
} 
\end{tcolorbox}
\captionof{figure}{The task details of the SayCan task and its task plan learned by Vicuna}
\label{program:saycan_Vicuna}
\end{figure*} \FloatBarrier
\section{The learned task plan by ChatGPT}
\label{appendix.gpt_4}
 \begin{center}
\begin{tcolorbox}[enhanced,title=Task 1,
attach boxed title to top left=
{xshift=2mm,yshift=-2mm},
boxed title style={size=small,colback=black},colback=black!-15!white,label={khan_task1}]
\textbf{Task Definition}: Fraction division in simplest form (dataset size:199).\\\\\textbf{Example}: Reduce to lowest terms: $\frac{1}{9} \div \frac{7}{6}$\\\\
\textbf{Learned natural language plan $p$ (zero-shot cot)}:
\\\\\small{To divide fractions, we can flip the second fraction and multiply. Then, we can simplify the resulting fraction by dividing the numerator and denominator by their greatest common factor. When dividing fractions, we may need to flip and multiply more than once.
\\\\Solution 1:To divide two fractions, we can flip the second fraction and multiply. Then, we can simplify the resulting fraction by dividing the numerator and denominator by their greatest common factor.
\\\\Solution 2: Another way to divide fractions is to convert them into decimals and then divide. This can be useful when dealing with fractions that have large denominators.
\\\\Solution 3: When dividing mixed numbers, we can convert them into improper fractions and then apply the same rules for dividing fractions.
\\\\Solution 4: In some cases, we may need to simplify the fractions before dividing. We can do this by finding the prime factors of the numerator and denominator and canceling out any common factors.
\\\\Solution 5: When dividing fractions with variables, we can use the same rules as dividing numerical fractions, but we need to be careful with the algebraic manipulation and ensure that we do not divide by zero.} 
\\\\
\normalsize\textbf{Learned natural language plan $p$ (few-shot cot)}:
\\\\\small{To divide fractions, we can either multiply the first fraction by the reciprocal of the second fraction or invert the second fraction and multiply.
\\\\
Similar solutions involve dividing two fractions by multiplying the first fraction with the reciprocal of the second fraction. The resulting fractions are then simplified 
by dividing both the numerator and denominator by their greatest common factor. 
}
\end{tcolorbox}
\label{fig:task1}
\captionof{figure}{The task details of task 1 of the AMPS dataset and its learned task plan}
\end{center} 

 \begin{figure}
\begin{tcolorbox}[enhanced,title=Task 2,
attach boxed title to top left=
{xshift=2mm,yshift=-2mm},
boxed title style={size=small,colback=black},colback=black!-15!white]
\textbf{Task Definition}: Calculating the value of a composite function (dataset size:394).\\\\\textbf{Example}: $h(n) = 2n$, $f(x) = 2x^{2}+6x-h(x)$, $ f(h(-2)) = {?} $\\\\
\textbf{Learned natural language plan $p$ (zero-shot cot)}:
\\\\\small{Solution 1: When substituting a value into a function that is defined in terms of another function, make sure to substitute the entire expression for the second function, not just the value of the variable.
\\\\
Solution 2: When evaluating a function with a nested function inside, substitute the value of the nested function into the equation for the outer function, not just the variable of the nested function.
\\\\
Solution 3: Remember to use the correct order of operations when evaluating a function, especially when there are multiple terms and exponents involved.
\\\\
Solution 4: Double-check your calculations and make sure to use the correct signs and coefficients when simplifying expressions.
\\\\
Solution 5: Practice and review basic algebraic principles, such as the distributive property and combining like terms, to help simplify complex functions.} 
\\\\
\textbf{Learned natural language plan $p$ (few-shot cot)}:
\\\\\small{Two solutions are provided for finding $f(h(-2))$ and $h(f(2))$, one involving substitution and one involving evaluation of the expressions. \\\\To find $h(g(-8))$ and $h(f(-1))$, one can substitute the given value into the expression for $g(n)$ or $f(n)$ and then substitute the resulting value into the expression for $h(x)$. \\\\Two solutions are also provided for finding $f(g(-3))$ and $g(f(-4))$, again involving substitution and evaluation of the expressions.
}

\end{tcolorbox}
\captionof{figure}{The task details of task 2 of the AMPS dataset and its learned task plan}
\label{program:task2}
\end{figure} 

 \begin{figure*}[h]
\begin{tcolorbox}[enhanced,title=Task 3,
attach boxed title to top left=
{xshift=2mm,yshift=-2mm},
boxed title style={size=small,colback=black},colback=black!-15!white]
\textbf{Task Definition}: Angle Calculation (dataset size: 368).\\\\\textbf{Example}: If $\angle AOC = 180^\circ$ and $\angle BOC = 160^\circ$, what is $\angle AOB$, in degrees? \\\\
\textbf{Learned natural language plan $p$ (zero-shot cot)}:
\\\\\small{Here are the five main solutions for finding unknown angles in triangles:
\\\\
1. Use the fact that the sum of angles in a triangle is 180 degrees. Subtract the known angles from 180 degrees to find the unknown angle.
\\\\
2. Use the Law of Cosines to find the unknown angle in a triangle with two sides and the included angle known.
\\\\
3. Use the Pythagorean Theorem and trigonometric functions to find the unknown angle in a right triangle.
\\\\
4. Use the fact that the sum of all angles in a triangle is 180 degrees to find the third angle when two angles are known.
\\\\
5. Use the Law of Sines and trigonometric functions to find the unknown angle in a triangle with one side and two angles known.
} 
\\\\
\textbf{Learned natural language plan $p$ (few-shot cot)}:
\\\\\small{There are two similar solutions. The first solution subtracts the measure of $\angle RPS$ from $180^\circ$ to find $m \angle QPR$. The second solution uses the fact that the sum of angles in a triangle is $180^\circ$, but there is an error in the calculation that is corrected by subtracting $m \angle RPS$ from $180^\circ$.\\\\
There are two solutions to find $m \angle MON$: using the fact that the sum of angles in a triangle is $180^\circ$ and the fact that $m \angle LOM$ and $m \angle MON$ are supplementary angles. The first solution yields $m \angle MON = 175^\circ$ while the second solution yields $m \angle MON \approx 175^\circ$ by using trigonometry.\\\\
There are two solutions given to find the measure of angle LOM. The first solution uses the fact that angles LOM and MON are supplementary, while the second solution uses the Law of Cosines. Both solutions yield the answer of 56 degrees.
\\\\
Solution: To find the measure of an angle, we can use the Law of Sines to find the length of the opposite side and then use the Law of Cosines to find the measure of the angle. This method can be applied to different situations where we need to find the measure of an angle, such as in triangles or other geometric shapes.
}
\end{tcolorbox}
\captionof{figure}{The task details of task 3 of the AMPS dataset and its learned task plan}
\label{program:task3}
\end{figure*} \FloatBarrier

 \begin{figure*}[h]
\begin{tcolorbox}[enhanced,title=Task 4,
attach boxed title to top left=
{xshift=2mm,yshift=-2mm},
boxed title style={size=small,colback=black},colback=black!-15!white]
\textbf{Task Definition}: Calculating the length of the hypotenuse or a leg of the right triangle (dataset size: 94).\\\\\textbf{Example}: In the right triangle shown,  angle $A=30^\circ$ and $AC=18$. How long is $AB$? \\\\
\textbf{Learned natural language plan $p$ (zero-shot cot)}:
\\\\\small{ To find the length of a side in a right triangle with a 30 degree angle, there are several solutions. These include using the ratio of the side opposite the angle to the hypotenuse, the Pythagorean theorem, the special right triangle with angles of 30-60-90 degrees, trigonometric functions, and the Law of Sines or the Law of Cosines.\\\\
There are two main solutions to find the length of the side opposite a 30 degree angle in a right triangle. One solution involves using the ratio of the side opposite the angle to the hypotenuse, while another solution involves using the sine function. The first solution gives $BC = \frac{1}{2}AB$ for the first two questions and $BC = 5$ for the third question, while the second solution gives $BC = 2$ for the first question, $BC = 3$ for the second question, and $BC = \frac{\sqrt{3}}{2}$ for the third question.\\\\
To find the length of the side opposite a 30 degree angle in a right triangle, we can use the tangent function, which gives us a length of $3\sqrt{3}$. Alternatively, we can use the Law of Cosines to find the length, which gives us a length of approximately 6.19 units.\\\\
There are two solutions to find the length of the side opposite a 30 degree angle in a right triangle: using the sine function to get $AB = 2\sin(30^{\circ}) = 1$, or using the Pythagorean theorem and the special right triangle to get $AB = \frac{2\sqrt{6}}{3}$.\\\\
Two solutions for finding the length of the side opposite a 30 degree angle in a right triangle are using the cosine function, which gives us a length of $AC\cos(30^\circ)$, or using the Law of Sines to find the length of $AB$, which simplifies to $AB = 2AC\sin(30^\circ)$. Both solutions can be implemented in Python using the math library.\\\\
There are two ways to find the length of the side opposite a 30 degree angle in a right triangle. The first is to use the ratio of the side opposite the angle to the hypotenuse, which gives $AC = \frac{1}{2}AB\sqrt{3} = 6\sqrt{3}$. Another method is to use the Law of Sines, which gives $AC = 2AB\sin(30^{\circ}) = 12$.
} 
\\\\
\normalsize\textbf{Learned natural language plan $p$ (few-shot cot)}:
\\\\\small{There are several methods to find the length of a missing leg in a right triangle. 
\\\\The first solution is to use the ratio of sides in a 30-60-90 triangle, which is $1:\sqrt{3}:2$. \\\\The second solution is to use the Pythagorean theorem. Trigonometric functions such as sine and cosine can also be used. Furthermore, a Python algorithm can be implemented to solve for the missing leg in both situations. 
\\\\To find the length of a missing leg in a right triangle with a known angle and hypotenuse, we can use the sine or tangent function. \\\\Finally, there are two solutions to find the missing leg of a right triangle with a known angle of 60 degrees and hypotenuse, which involve using the cosine function or the Pythagorean theorem.
}
\end{tcolorbox}
\captionof{figure}{The task details of task 4 of the AMPS dataset and its learned task plan}
\label{program:task4}
\end{figure*} \FloatBarrier
 \begin{figure*}[h]
\begin{tcolorbox}[enhanced,title=Task 5,
attach boxed title to top left=
{xshift=2mm,yshift=-2mm},
boxed title style={size=small,colback=black},colback=black!-15!white]
\textbf{Task Definition}: Given a number A, find its factor/multiple in a list of numbers. (dataset size: 393).\\\\\textbf{Example}: Which of the following numbers is a factor of 56? $\{3,6,8,10,13\}$ \\\\
\textbf{Learned natural language plan $p$ (zero-shot cot)}:
\\\\\small{1. Using the modulo operator in Python.
\\\\
2. Checking the last digit of the number.
\\\\
3. Dividing the number by the factor and checking if the result is an integer.
\\\\
4. Using bitwise operators to check for powers of 2.
\\\\
5. Using the Euclidean algorithm to find the greatest common divisor.
\\\\
To identify multiples of a certain factor, we can use the modulo operator in Python. Another way is to check the last digit of the number. For example, to check if a number is a multiple of 8, we can check if its last three digits are divisible by 8 using the expression "$\textit{number} \% 1000 \% 8 == 0$". Other solutions include using loops, lists, and recursion to identify multiples of a certain factor.
} 
\\\\
\normalsize\textbf{Learned natural language plan $p$ (few-shot cot)}:
\\\\\small{There are several ways to check if a number is a multiple of another number. One way is to use the modulo operator or divide the number and check if the result is an integer.\\\\ Another way is to use prime factorization or the GCD/LCM formula. \\\\Additionally, we can use the Euclidean algorithm or the formula n = km. \\\\There are also methods that involve checking the sum of digits or using Python functions like "divmod()" and "all()".
}
\end{tcolorbox}
\captionof{figure}{The task details of task 5 of the AMPS dataset and its learned task plan}
\label{program:task5}
\end{figure*} \FloatBarrier

 \begin{figure*}[h]
\begin{tcolorbox}[enhanced,title=Task 6,
attach boxed title to top left=
{xshift=2mm,yshift=-2mm},
boxed title style={size=small,colback=black},colback=black!-15!white]
\textbf{Task Definition}: Calculating the significant figures in a number (dataset size: 197).\\\\\textbf{Example}: How many significant figures does $00.033$ have? \\\\
\textbf{Learned natural language plan $p$ (zero-shot cot)}:
\\\\\small{To determine the number of significant figures, count all non-zero digits and any zeros between non-zero digits. Therefore, the number 0197.01239440 has 11 significant figures, the number 0798 has 3 significant figures, and the number 0.03 has 1 significant figure. Leading zeros are not significant. \\\\Another way to determine the number of significant figures is to look at the precision of the measuring instrument. It is also important to consider the rules for rounding when performing calculations with significant figures.
\\\\Solution: To determine the number of significant figures, count all non-zero digits and any zeros between non-zero digits. Ignore leading zeros. If a number ends in zeros, the zeros are only significant if there is a decimal point. For example, the number 0.0032 has 2 significant figures, while the number 3200 has 2 significant figures. Additionally, the number 5000 has 1 significant figure, while the number 5000. has 4 significant figures.
\\\\The solutions provided all involve counting the significant figures in a number by counting non-zero digits and zeros between non-zero digits. One specific Python algorithm involves converting the number to a string, removing leading zeros, trailing zeros after a decimal point, and then returning the length of the resulting string as the number of significant figures. Other solutions may involve using scientific notation or logarithms to determine significant figures. 
} 
\\\\
\normalsize\textbf{Learned natural language plan $p$ (few-shot cot)}:
\\\\\small{Solution: To determine the number of significant figures in a number, count all non-zero digits and any zeros between them. Trailing zeros after the last non-zero digit are only significant if there is a decimal point present. Therefore, $0.253971000$ has 9 significant figures and $292.785854000$ has 12 significant figures.
\\\\To determine the number of significant figures in a number, one solution is to count all non-zero digits and any zeros between them, including trailing zeros after the last non-zero digit if there is a decimal point. Another solution is a Python algorithm that converts the number to a string and counts the number of characters excluding leading and trailing zeros. This would output the number of significant figures.}
\end{tcolorbox}
\captionof{figure}{The task details of task 6 of the AMPS dataset and its learned task plan}
\label{program:task6}
\end{figure*} \FloatBarrier

 \begin{figure*}[h]
\begin{tcolorbox}[enhanced,title=Task 7,
attach boxed title to top left=
{xshift=2mm,yshift=-2mm},
boxed title style={size=small,colback=black},colback=black!-15!white]
\textbf{Task Definition}: Calculating the sin/cos value of an angle based on the length of legs and other conditions in a triangle. (dataset size: 97).\\\\\textbf{Example}: $\overline{AC}$ is $8$ units long, $\overline{BC}$ is $15$ units long, $\overline{AB}$ is $17$ units long ,What is $\cos(\angle ABC)$ ? \\\\
\textbf{Learned natural language plan $p$ (zero-shot cot)}:
\\\\\small{There are two solutions for finding the value of an angle in a triangle. Solution 1 involves using the Law of Cosines to calculate $\cos(\angle ABC)$. \\\\Solution 2 involves using the Pythagorean Theorem to find the length of the third side of the triangle and then using the definition of sine to calculate $\sin(\angle ABC)$. To find the length of the hypotenuse of a right triangle using the Pythagorean Theorem, use the formula $(\overline{AC})^2 + (\overline{BC})^2 = (\overline{AB})^2$. python functions can also be used to calculate these values.
} 
\\\\
\textbf{Learned natural language plan $p$ (few-shot cot)}:
\\\\\small{When working with trigonometric functions, it is essential to identify the correct sides of the triangle. Labeling the angles and sides of the triangle can help avoid confusion. In addition, simplifying fractions and using the appropriate trigonometric function for the given situation are crucial for obtaining the correct answer.
}
\end{tcolorbox}
\captionof{figure}{The task details of task 7 and its learned task plan}
\label{fig:task7}
\end{figure*} \FloatBarrier

 \begin{figure*}[h]
\begin{tcolorbox}[enhanced,title=Task 8,
attach boxed title to top left=
{xshift=2mm,yshift=-2mm},
boxed title style={size=small,colback=black},colback=black!-15!white]
\textbf{Task Definition}: Fraction Division. (dataset size: 192).\\\\\textbf{Example}: $\frac{1}{8} \div \frac{8}{3} = {?}$ \\\\
\textbf{Learned natural language plan $p$ (zero-shot cot)}:
\\\\\small{Here are the main solutions for dividing fractions:\\
1. Flip the second fraction and multiply.\\
For example, to divide $\dfrac{2}{7}$ by $\dfrac{8}{3}$, we flip $\dfrac{8}{3}$ to get $\dfrac{3}{8}$ and multiply to get 
$\dfrac{2}{7} \times \dfrac{3}{8} = \dfrac{6}{56}$.
\\\\
2. Convert mixed numbers to improper fractions, flip the second fraction, and multiply.
\\For example, to divide $2 \dfrac{1}{2}$ by $\dfrac{3}{4}$, we convert $2 \dfrac{1}{2}$ to $\dfrac{5}{2}$, flip $\dfrac{3}{4}$ to get
 $\dfrac{4}{3}$, and multiply to get $\dfrac{5}{2} \times \dfrac{4}{3} = \dfrac{10}{3}$.
\\\\
3. Use cross-multiplication.\\ 
For example, to divide $\dfrac{2}{5}$ by $\dfrac{1}{3}$, we cross-multiply to get $2 \times 3 = 5x$, which simplifies to 
$x = \dfrac{6}{5}$.
\\\\
4. Use the reciprocal of the second fraction.\\
 For example, to divide $\dfrac{3}{4}$ by $\dfrac{5}{6}$, we take the reciprocal of $\dfrac{5}{6}$ to get $\dfrac{6}{5}$ and 
multiply to get $\dfrac{3}{4} \times \dfrac{6}{5} = \dfrac{18}{20}$.
\\\\
5. Divide both fractions by their greatest common factor.\\
 For example, to divide $\dfrac{16}{24}$ by $\dfrac{4}{9}$, we simplify both fractions by dividing by their greatest common factor of 8 to get $\dfrac{2}{3}$ and $\dfrac{1}{2}$, respectively. Then, we flip $\dfrac{1}{2}$ and multiply to get $\dfrac{2}{3} \times \dfrac{2}{1} = \dfrac{4}{3}$.
} 
\\\\
\normalsize\textbf{Learned natural language plan $p$ (few-shot cot)}:
\\\\\small{There are two main solutions for dividing fractions: inverting the second fraction and multiplying, or converting the division sign to a multiplication sign and flipping the second fraction. Both methods yield the same result and can be used to solve all three questions. Additionally, Python code can be used to solve each question by defining the numerators and denominators of each fraction and then dividing them. \\\\Another solution involves simplifying the fractions by finding a common factor, then dividing their numerators. Finally, the fractions module in Python can be used to simplify the result to a fraction.
}
\end{tcolorbox}
\captionof{figure}{The task details of task 8 of the AMPS dataset and its learned task plan}
\label{program:task8}
\end{figure*} \FloatBarrier

 \begin{figure*}[h]
\begin{tcolorbox}[enhanced,title=Task 9,
attach boxed title to top left=
{xshift=2mm,yshift=-2mm},
boxed title style={size=small,colback=black},colback=black!-15!white]
\textbf{Task Definition}: Fraction reduction in the form of the mixed number (dataset size: 400).\\\\\textbf{Example}: Express your answer as a mixed number simplified to lowest terms. $9\dfrac{3}{4}-5\dfrac{4}{6} = {?}$ \\\\
\textbf{Learned natural language plan $p$ (zero-shot cot)}:
\\\\\small{
 To add or subtract mixed numbers, we first convert them to improper fractions, find a common denominator, add or subtract the whole numbers separately, and simplify the resulting fraction if possible. If the final fraction is improper, we can convert it back to a mixed number. This method applies to all situations involving mixed numbers.
} 
\\\\
\normalsize\textbf{Learned natural language plan $p$ (few-shot cot)}:
\\\\\small{One solution for subtracting mixed numbers with fractions is to convert the mixed numbers to improper fractions, find a common denominator, subtract the fractions, and simplify the resulting improper fraction to a mixed number.\\\\ Another solution is to convert the mixed numbers to improper fractions, subtract the fractions, and simplify the resulting improper fraction to a mixed number without finding a common denominator. Both solutions are presented in Python algorithms.
}
\end{tcolorbox}
\captionof{figure}{The task details of task 9 of the AMPS dataset and its learned task plan}
\label{program:task9}
\end{figure*} \FloatBarrier

 \begin{figure*}[h]
\begin{tcolorbox}[enhanced,title=Task 10,
attach boxed title to top left=
{xshift=2mm,yshift=-2mm},
boxed title style={size=small,colback=black},colback=black!-15!white]
\textbf{Task Definition}: Calculating the length of a leg in a triangle based on the length of another leg and the sin, cos, and tan value of an angle. (dataset size: 593).\\\\\textbf{Example}: $\overline{BC}=9$, $\sin(\angle ABC )=\frac{ \sqrt{2}}{2}$,$\cos( \angle ABC )=\frac{ \sqrt{2}}{2}$, $\tan( \angle ABC )=1$, then $\overline{AC}={?}$ \\\\
\textbf{Learned natural language plan $p$ (zero-shot cot)}:
\\\\\small{
Solution 1: For both questions, we can use trigonometric ratios involving the given angle to find the length of one side of the triangle. Then, we can use the Pythagorean theorem and simplify to find the length of the other side.
\\\\
Solution 2: Another method is to use the fact that the sum of the squares of the sine and cosine of an angle is equal to 1. We can substitute in the values we know and solve for the length of the desired side.
\\\\
Solution 3: In the second question, we can also use the law of cosines to find the length of one side of the triangle, and then use the Pythagorean theorem to find the length of the other side.
\\\\
Solution 4: If we are given the area of the triangle, we can use the formula $A = 1/2 \cdot ab \cdot \sin(C)$ to find the length of one side, and then use the Pythagorean theorem to find the length of the other side.
\\\\
Solution 5: If we are given the radius of the circumcircle of the triangle, we can use the formula $R = a \cdot b \cdot c / 4A$ to find the length of one side, and then use the Pythagorean theorem to find the length of the other side.
} 
\\\\
\normalsize\textbf{Learned natural language plan $p$ (few-shot cot)}:
\\\\\small{There are two main solutions to find the missing side length in the given right triangle.\\\\ Solution 1 involves using the sine function to calculate the length of the missing side.\\\\ Solution 2 involves using the Pythagorean theorem. For each question, we can choose either solution depending on the given information.
}
\end{tcolorbox}
\captionof{figure}{The task details of task 10 of the AMPS dataset and its learned task plan}
\label{program:task10}
\end{figure*} \FloatBarrier

 \begin{figure*}[h]
\begin{tcolorbox}[enhanced,title=Causal Judgment Task,
attach boxed title to top left=
{xshift=2mm,yshift=-2mm},
boxed title style={size=small,colback=black},colback=black!-15!white]
\textbf{Task Definition}: Given a short story, LLMs need to
judge if event X causes event Y.  \\\\\textbf{Example}: The receptionist in the philosophy department keeps her desk stocked with pens. The administrative assistants are allowed to take the pens, but faculty members are supposed to buy their own. The administrative assistants typically do take the pens. Unfortunately, so do the faculty members. The receptionist has repeatedly emailed them reminders that only administrative assistants are allowed to take the pens. On Monday morning, one of the administrative assistants encounters Professor Smith walking past the receptionist's desk. Both take pens. Later that day, the receptionist needs to take an important message... but she has a problem. There are no pens left on her desk. 

Did the administrative assistant cause the problem?
\\\\
\textbf{Learned natural language plan $p$ (zero-shot cot)}:
\\\\\small{
Consider the agent's intentions and actions to determine causality.
Use counterfactual scenarios to determine causality.
Consider normative expectations and social norms to determine causality.
Consider the agent's intentions and actions to determine causality.
Use counterfactual scenarios to determine causality.
Social norms and normative expectations can play a role in determining causality.

Use counterfactual scenarios to determine causality by imagining what would have happened if the causal factor had not occurred. Consider social norms and normative expectations as they can play a role in determining causality, as actions that violate these norms may be seen as causing negative outcomes. 
} 
\\\\
\normalsize\textbf{Learned natural language plan $p$ (few-shot cot)}:
\\\\\small{To determine causality, consider the counterfactual scenario, temporal order, potential mechanisms, and control for other factors. Evaluate the strength of the relationship and plausibility of the mechanism, while considering alternative explanations for the outcome.
}
\end{tcolorbox}
\captionof{figure}{The task details of the Causal Judgment task and its learned task plan}
\label{program:judge}
\end{figure*} \FloatBarrier

 \begin{figure*}[h]
\begin{tcolorbox}[enhanced,title=Counterfactual Reasoning Assessment Task,
attach boxed title to top left=
{xshift=2mm,yshift=-2mm},
boxed title style={size=small,colback=black},colback=black!-15!white]
\textbf{Task Definition}: Given a premise, LLMs need to
judge the possible result if the event
in the premise does not happen. \\\\\textbf{Example}: Premise: A woman does not increase his own capacity. 

Question: What would have happened if she had increased his own capacity?

Options: (A) The woman would have felt accomplished. (B) The woman would not have felt accomplished. (C) That is not possible.
\\\\
\textbf{Learned natural language plan $p$ (zero-shot cot)}:
\\\\\small{
When answering inference questions, remember that the scenario presented is hypothetical and did not actually happen. Choose the most logical answer based on cause-and-effect relationships.

Avoid making assumptions about cause-and-effect relationships in anti-causal inference questions. Focus on the information provided in the premise and choose the most reasonable answer.

Summarize similar solutions into one rule and maintain solutions for solving different situations. When generating a solution, write no more than two sentences.

Focus on the information provided in the premise and avoid making assumptions about cause-and-effect relationships.

For anti-causal inference questions, choose the most logical answer based on cause-and-effect relationships.

Acknowledge that the premise does not provide enough information to determine a cause-and-effect relationship when answering anti-causal inference questions. Choose the most reasonable answer based on the given scenario.

Focus on the information provided in the premise and avoid making assumptions about cause-and-effect relationships.

Choose the most logical answer based on the given scenario for non-causal inference questions.
For anti-causal inference questions, acknowledge that the premise does not provide enough information to determine a cause-and-effect relationship and choose the most reasonable answer based on the given scenario.
} 
\\\\
\normalsize\textbf{Learned natural language plan $p$ (few-shot cot)}:
\\\\\small{Consider all possible outcomes and analyze the situation to determine the likelihood of each option.

Use logic and reasoning to determine the most likely outcome by considering all possible scenarios and eliminating unlikely or impossible options.

Focus on the cause and effect relationship between the premise and the question to determine the most likely outcome and avoid errors in reasoning.

Carefully analyze the premise and question to determine all possible outcomes and their likelihood to avoid errors in reasoning.

Use logic and reasoning to consider all possible scenarios and eliminate unlikely or impossible options when answering anti-casual inference questions.

Focus on the cause and effect relationship between the premise and the question to determine the most likely outcome and avoid errors in reasoning.
}
\end{tcolorbox}
\captionof{figure}{The task details of the Counterfactual Reasoning Assessment task and its learned task plan}
\label{program:crass}
\end{figure*} \FloatBarrier

 \begin{figure*}[h]
\begin{tcolorbox}[enhanced,title=Logical Reasoning (LSAT) Task,
attach boxed title to top left=
{xshift=2mm,yshift=-2mm},
boxed title style={size=small,colback=black},colback=black!-15!white]
\textbf{Task Definition}: Given a passage and a question,
LLMs need to infer which option
is true. \\\\\textbf{Example}: Passage: Hana said she was not going to invite her brothers to her birthday party. However, among the gifts Hana received at her party was a recording in which she had expressed an interest. Since her brothers had planned to give her that recording, at least some of Hana's brothers must have been among the guests at Hana's birthday party after all.

Question: A reasoning error in the argument is that the argument

Options: (A) disregards the possibility that a change of mind might be justified by a change in circumstances (B)treats the fact of someone's presence at a given event as a guarantee that that person had a legitimate reason to be at that event (C)uses a term that is intrinsically evaluative as though that term was purely descriptive (D)fails to establish that something true of some people is true of only those people (E)overlooks the possibility that a person's interest in one kind of thing is compatible with that person's interest in a different kind of thing.
\\\\
\textbf{Learned natural language plan $p$ (zero-shot cot)}:
\\\\\small{
Identify the conclusion and premises of an argument and look for any assumptions that the argument relies on.
Consider any counterexamples or objections that could weaken the argument.
Look for evidence or reasoning that supports or undermines the conclusion. 

Identify the conclusion and premises of the argument, and look for any assumptions it relies on. 
Consider any counter examples or objections that could weaken the argument. 
Look for evidence or reasoning that supports or undermines the conclusion. 
} 
\\\\
\normalsize\textbf{Learned natural language plan $p$ (few-shot cot)}:
\\\\\small{Look for an answer choice that resolves the paradox by providing a possible explanation for why the observed phenomenon occurred.
Pay attention to the language used in the passage, including any conditional statements or cause-and-effect relationships.
Consider the context of the passage and any relevant background information that may help to explain the observed phenomenon.
}
\end{tcolorbox}
\captionof{figure}{The task details of the Logical Reasoning (LSAT) task and its learned task plan}
\label{program:lsat}
\end{figure*} \FloatBarrier
 \begin{figure*}[h]
\begin{tcolorbox}[enhanced,title=Date Understanding Task,
attach boxed title to top left=
{xshift=2mm,yshift=-2mm},
boxed title style={size=small,colback=black},colback=black!-15!white]
\textbf{Task Definition}: LLMs need to infer the date based
on the context. \\\\\textbf{Example}: The deadline is Jun 1, 2021, which is 2 days away from now. What is the date one year ago from today in MM/DD/YYYY?\\\\
\textbf{Learned natural language plan $p$ (zero-shot cot)}:
\\\\\small{
To find a date a year ago, subtract 1 from the year and keep the same month and day. To find a date a month ago, subtract 1 from the month and keep the same day and year. To find a date 24 hours later, add 1 to the day while keeping the same month and year.

To find a date a year ago, subtract 1 from the year and keep the same month and day. To find a date a month ago, subtract 1 from the month and keep the same day and year. To find a date 24 hours later, add 1 to the day while keeping the same month and year.

To find a date a year ago, subtract 1 from the year and keep the same month and day. To find a date a month ago, subtract 1 from the month and keep the same day and year. To find a date 24 hours later, add 1 to the day while keeping the same month and year.
} 
\\\\
\normalsize\textbf{Learned natural language plan $p$ (few-shot cot)}:
\\\\\small{Always double-check the date and format when converting between different date formats. Use a datetime module or calculator to ensure accuracy.

When dealing with holidays or special events, double-check the year to ensure you're referencing the correct date. Use a calendar or reference guide to confirm the date.

Always use a reliable source or tool to ensure accuracy when dealing with dates. Double-check the date format and any conversions made to avoid errors. When referencing holidays or special events, double-check the year using a calendar or reference guide to confirm the date and avoid mistakes. Keep separate solutions for different situations and limit each solution to no more than two sentences.

Always double-check dates and formats using reliable sources or tools to ensure accuracy.

When dealing with holidays or special events, double-check the year and use a calendar or reference guide to confirm the date and avoid mistakes. 

Note: Maintain separate solutions for different situations and limit each solution to no more than two sentences.
}
\end{tcolorbox}
\captionof{figure}{The task details of the date understanding task and its learned task plan}
\label{program:date}
\end{figure*} \FloatBarrier

 \begin{figure*}[h]
\begin{tcolorbox}[enhanced,title=Last Letter Concatenation Task,
attach boxed title to top left=
{xshift=2mm,yshift=-2mm},
boxed title style={size=small,colback=black},colback=black!-15!white]
\textbf{Task Definition}: Given n words, LLMs need to extract and concatenate the last letters
of the n words. \\\\\textbf{Example}: Take the last letters of each words in "spur drowning Japan dialect valet " and concatenate them \\\\
\textbf{Learned natural language plan $p$ (zero-shot cot)}:
\\\\\small{
To find the last letters of each word in a phrase, identify the last letter of each word and write them down in order. Then, concatenate the letters to form the final answer. Make sure to follow this process for each phrase, as different phrases may require different solutions.
} 
\\\\
\normalsize\textbf{Learned natural language plan $p$ (few-shot cot)}:
\\\\\small{To concatenate the last letters of words in a given phrase, first write down the words and identify their last letters. Then, concatenate the last letters in the order they appear in the phrase to get the final answer. Double-check your answer to avoid errors. 

To concatenate the last letters of words in a given phrase, first list the words and identify their last letters. Then, concatenate the last letters in the order they appear in the phrase to get the final answer. Double-checking your answer can help avoid errors. It's important to have separate solutions for different situations and to keep each solution concise, no more than two sentences.

To concatenate the last letters of words in a given phrase, list the words and identify their last letters. Then, concatenate the last letters in the order they appear in the phrase to get the final answer. Always double-check your answer for accuracy.
}
\end{tcolorbox}
\captionof{figure}{The task details of the Last Letter Concatenation task and its learned task plan}
\label{program:last_letter}
\end{figure*} \FloatBarrier

 \begin{figure*}[h]
\begin{tcolorbox}[enhanced,title=SayCan Task,
attach boxed title to top left=
{xshift=2mm,yshift=-2mm},
boxed title style={size=small,colback=black},colback=black!-15!white]
\textbf{Task Definition}: Given an instruction, LLMs need to
generate an action sequence for the
robot to complete the instruction. \\\\\textbf{Example}: How would you move the chips bag from the table to
the counter?
\\\\
\textbf{Learned natural language plan $p$ (zero-shot cot)}:
\\\\\small{
Identify the requested item, locate it in the designated area, and bring it to the user.

Gather all items on a table and dispose of them properly, confirming completion with the user.

Identify the specific type of item requested, search for it in the designated area, and bring it to the user. 
} 
\\\\
\normalsize\textbf{Learned natural language plan $p$ (few-shot cot)}:
\\\\\small{Solution 1: Before preparing a meal, make a list of necessary ingredients and consider any dietary restrictions or preferences. Follow a recipe or create a dish based on the chosen ingredients.

Solution 2: When cleaning a space, start by decluttering and organizing items that are out of place. Then, dust surfaces and sweep or vacuum the floors for a clean and tidy environment.

Solution 3: Before packing for a trip, make a list of necessary items and consider the length of the trip, weather, and planned activities. Pack clothing and other essentials based on these factors to ensure a comfortable and enjoyable trip.
}
\end{tcolorbox}
\captionof{figure}{The task details of the SayCan task and its learned task plan}
\label{program:saycan}
\end{figure*} \FloatBarrier

\begin{figure*}
\tcbset{colback=red!-15!white,colframe=red!65!yellow,fonttitle=\bfseries}
\begin{tcolorbox}[breakable,title=Continual update process of the task plan for calculating angle task, sidebyside align=top]
\small\textbf{Question}:If $\angle QPS = 180^\circ$ and $\angle QPR = 134^\circ$, what is $\angle RPS$, in degrees?\\
\\\\
\textbf{Task plan learned by our method (currently in training)}:\\\small{Use the Law of Sines or Law of Cosines to find unknown angles in a triangle.
\\Another solution is to use the Pythagorean Theorem to find the length of the hypotenuse or one of the legs in a right triangle.\\
\textbf{Solution guided by the plan}: We need at least one more angle measurement or side length to use trigonometric ratios or the Law of Cosines to solve for the missing angle.}
\\\\
\textbf{Task plan learned by our method (after the training)}:\\\small{1. Use the fact that the sum of angles in a triangle is 180 degrees. Subtract the known angles from 180 degrees to find the unknown angle.\\
2. Use the Law of Cosines to find the unknown angle in a triangle with two sides and the included angle known.\\
3. Use the Pythagorean Theorem and trigonometric functions to find the unknown angle in a right triangle.\\4. Use the fact that the sum of all angles in a triangle is 180 degrees to find the third angle when two angles are known.
\\5. Use the Law of Sines and trigonometric functions to find the unknown angle in a triangle with one side and two angles known.
\\\textbf{Solution guided by the plan}: we can use solution 1. Since $\angle QPS = 180^\circ$, we know that $\angle QPR$ and $\angle RPS$ are adjacent angles that add up to $180^\circ$. \\Therefore, we can subtract $\angle QPR$ from $180^\circ$ to find $\angle RPS$:
\\
$$\angle RPS = 180^\circ - \angle QPR =46^\circ$$
Therefore, $\angle RPS$ is $46^\circ$}

\end{tcolorbox}
\captionof{figure}{Task plans and solutions guided by them in different learning phases. 'Currently in training means that the training process is starting but has not been finished.}
\label{analysis:continual}
\vspace{-5pt}
\end{figure*}
\FloatBarrier
\section{Case Study}
\label{appendix.case}
 \begin{figure*}
\begin{tcolorbox}[enhanced,title=Task plan for the task 1,
attach boxed title to top left=
{xshift=2mm,yshift=-2mm},
boxed title style={size=small,colback=black},colback=black!-15!white,label={khan_task1}]
\textbf{Task Definition}: Fraction division in simplest form (dataset size:199).\\\\\textbf{Example}: Reduce to lowest terms: $\frac{1}{9} \div \frac{7}{6}$\\\\
\textbf{Natural language task plan $p$ by self plan method }:
\small{\\\\Solution 1:  Invert the second fraction and multiply.
\\\\Solution 2: You can use the fact that division by a fraction is equivalent to multiplication by its reciprocal. Then, simplify by dividing both the numerator and denominator by their Least common multiple. }
\\\\
\normalsize\textbf{Natural language task plan $p$ by self plan with selection method }:
\small{\\\\
Solution 1: To reduce a fraction to its lowest terms, find the greatest common factor (GCF) of the numerator and denominator, then divide both by the GCF. Repeat until the GCF is 1. \\\\
Solution 2: Use the Euclidean algorithm by dividing the numerator by the denominator, and taking the remainder. Then divide the denominator by the remainder and take the remainder again. Repeat until the remainder is 0, and the last divisor is the GCF. Divide both numerator and denominator by the GCF to get the lowest terms. \\\\Solution 3: Express the fraction as a product of its prime factors, then cancel any common factors between the numerator and denominator.}
\\\\
\normalsize\textbf{Learned natural language task plan $p$ by learning to plan method (zero-shot cot)}:
\\\\\small{To divide fractions, we can flip the second fraction and multiply. Then, we can simplify the resulting fraction by dividing the numerator and denominator by their greatest common factor. When dividing fractions, we may need to flip and multiply more than once.
\\\\Solution 1:To divide two fractions, we can flip the second fraction and multiply. Then, we can simplify the resulting fraction by dividing the numerator and denominator by their greatest common factor.
\\\\Solution 2: Another way to divide fractions is to convert them into decimals and then divide. This can be useful when dealing with fractions that have large denominators.
\\\\Solution 3: When dividing mixed numbers, we can convert them into improper fractions and then apply the same rules for dividing fractions.
\\\\Solution 4: In some cases, we may need to simplify the fractions before dividing. We can do this by finding the prime factors of the numerator and denominator and canceling out any common factors.
\\\\Solution 5: When dividing fractions with variables, we can use the same rules as dividing numerical fractions, but we need to be careful with the algebraic manipulation and ensure that we do not divide by zero.} 
\\\\
\end{tcolorbox}
\captionof{figure}{Task plans for the task 1 of the AMPS dataset}
\label{fig: plans}
\end{figure*}
\end{document}